# Beyond Major Product Prediction: Reproducing Reaction Mechanisms with Machine Learning Models Trained on a Large-Scale Mechanistic Dataset

Joonyoung F. Joung[1], Mun Hong Fong[1], Jihye Roh[1], Zhengkai Tu[2], John Bradshaw[1], and Connor W. Coley[1, 2, *]

[1]Department of Chemical Engineering, Massachusetts Institute of Technology, Cambridge, Massachusetts 02139, United States

[2]Department of Electrical Engineering and Computer Science, Massachusetts Institute of Technology, Cambridge, Massachusetts 02139, United States

*E-mail: ccoley@mit.edu (C.W.C.)

**Abstract**

Mechanistic understanding of organic reactions can facilitate reaction development, impurity prediction, and in principle, reaction discovery. While several machine learning models have sought to address the task of predicting reaction products, their extension to predicting reaction mechanisms has been impeded by the lack of a corresponding mechanistic dataset. In this study, we construct such a dataset by imputing intermediates between experimentally reported reactants and products using expert reaction templates and train several machine learning models on the resulting dataset of 5,184,184 elementary steps. We explore the performance and capabilities of these models, focusing on their ability to predict reaction pathways and recapitulate the roles of catalysts and reagents. Additionally, we demonstrate the potential of mechanistic models in predicting impurities, often overlooked by conventional models. We conclude by evaluating the generalizability of mechanistic models to new reaction types, revealing challenges related to dataset diversity, consecutive predictions, and violations of atom conservation.

**Introduction**

Anticipating the outcomes of chemical reactions given specific reactants and conditions remains a formidable challenge in the field of chemistry. Expert chemists rely on a comprehension of reaction mechanisms as a guiding framework for predicting the probable outcomes of reactions, often recorded as "arrow pushing" diagrams. However, discerning a plausible mechanism through which a reaction proceeds by intuition alone is not always straightforward, resulting in development of computational methods and quantitative analyses for their investigation.

First principles quantum chemical calculations offer one approach to the proposal and validation of reaction mechanisms when provided with reactants and known products. Despite the development of techniques such as growing string[1] and nudged elastic band[2] to uncover elementary reaction pathways and transition states connecting initial and final states, these methods are computationally expensive and reliant on expertise to predefine hypothetical pathways for evaluation. Studies focused on reaction mechanisms using quantum chemical calculations often encounter limitations, primarily restricting their scope to reactions involving relatively small molecule sizes.[3]

If the goal of reaction outcome prediction is simplified to the prediction of major products, abstracting away details of the chemistry, the problem becomes amenable to data-driven solutions. A variety of machine learning models trained on experimental data reported in journal articles and patents have been applied to this task in recent years, leveraging problem formulations such as graph edit prediction with graph neural networks[4], machine translation of reactant SMILES strings[5] or encoded reactant graphs[6] into products, prediction of electron paths in the reaction[7], and classification of reaction templates.[8] While useful in many contexts, these end-to-end machine learning models often face criticism for their inability to *explain* the formation of products from given reactants in terminology consistent with how organic chemists explain reactivity.

In principle, machine learning models for product prediction could be retrained on mechanistic datasets to predict intermediate products as well. In practice, the lack of datasets describing "correct" (or at least widely agreed upon and plausible) reaction mechanisms has been a barrier. Various attempts have been made to address the lack of datasets containing reaction mechanisms. One such approach involves the construction of mechanistic datasets through quantum chemical calculations.[9] However, owing to computational expenses, datasets rooted in quantum mechanics tend to focus on reactions involving very small molecules, constrained to those with a maximum of seven or ten heavy atoms of C, N, and O.[9] While these datasets prove beneficial for developing novel machine learning models tailored to predict the transition state of specific reactions,[10] they are limited by their focus on small molecules and a few atom types to compromise computational costs.[11]

Large-scale datasets of chemical reactions extracted from the patent literature like USPTO-Full[12] and Pistachio[13], containing over 1 and 14 million reactions reported since 1976, respectively, serve as valuable resources for model training.[4-7, 8B, 8C, 14] While they do not contain any mechanistic information themselves, under strong assumptions, heuristics can be used to infer potential intermediates based on electronegativity. One noteworthy example of a model trained on these assumptions is ELECTRO,[7] which focuses on linear electron paths to model a chemical reaction as pairs of electrons moving along a single path through the reactant atoms. However, it is important to note that while the model infers electron movement from reactants and products, it is not grounded in expert-annotated mechanisms and may predict non-physical mechanisms, particularly overlooking the involvement of catalysts. Inferring reaction mechanisms from overall reactions through a minimal set of heuristics has inherent limitations, raising the necessity for curated datasets containing reaction mechanisms consistent with the field's understanding of reactivity.

In the pursuit of reaction mechanisms consistent with expert chemistry consensus, Jung and colleagues recently introduced an automated method for generating a mechanistic dataset, covering 94.8% of 33,000 USPTO reactions through the utilization of hand-coded mechanistic templates.[15] This dataset holds promise as a valuable source for machine learning applications. In addition to constructing precise mechanistic datasets, Baldi and coworkers have developed ReactionPredictor[16] and OrbChain[17] trained on a proprietary dataset containing widely accepted reaction mechanisms of 11k elementary reactions or 5.5k radical elementary reactions, respectively. Despite their ability to predict accurate mechanisms based on more authentic mechanistic datasets, these datasets may be narrower in terms of the scope of the reaction types that are covered than the other datasets such as USPTO-Full[12] and Pistachio[13].

In this work, following the general approach of Baldi, we introduce a new mechanistic dataset generated by applying expert-curated elementary reaction templates to impute the intermediates of a reaction dataset containing only reactants and products.[13] We do so, however, using experimentally-reported reactions derived from the patent literature to achieve a broader coverage of reaction types and reactant structures with relevant structural complexity. We identified the most popular 86 reaction types in Pistachio[13] and curated elementary reaction templates (Figure 1c) for each of these 86 reaction types with 175 different reaction conditions. By applying these expert elementary reaction templates to the reactants in Pistachio, we obtained the recorded products as well as unreported byproducts and side products. We systematically selected and preserved the mechanistic pathways leading to the formation of the recorded product for each entry, resulting in a comprehensive dataset comprising 1.3 million overall reactions and 5.8 million elementary reactions, thus constituting a large and authentic mechanistic dataset. We present three representative machine learning models, including graph-based,[4A] sequence-based,[5A] and graph-to-sequence based models,[6] trained to predict the next intermediate or product at the elementary reaction level, rather than the conventional approach of predicting only major products (Figure 1a). These models are designed to additionally predict byproducts and side products, such as the hydrogen bromide in the bromination using FeBr$_3$ shown in Figure 1a.

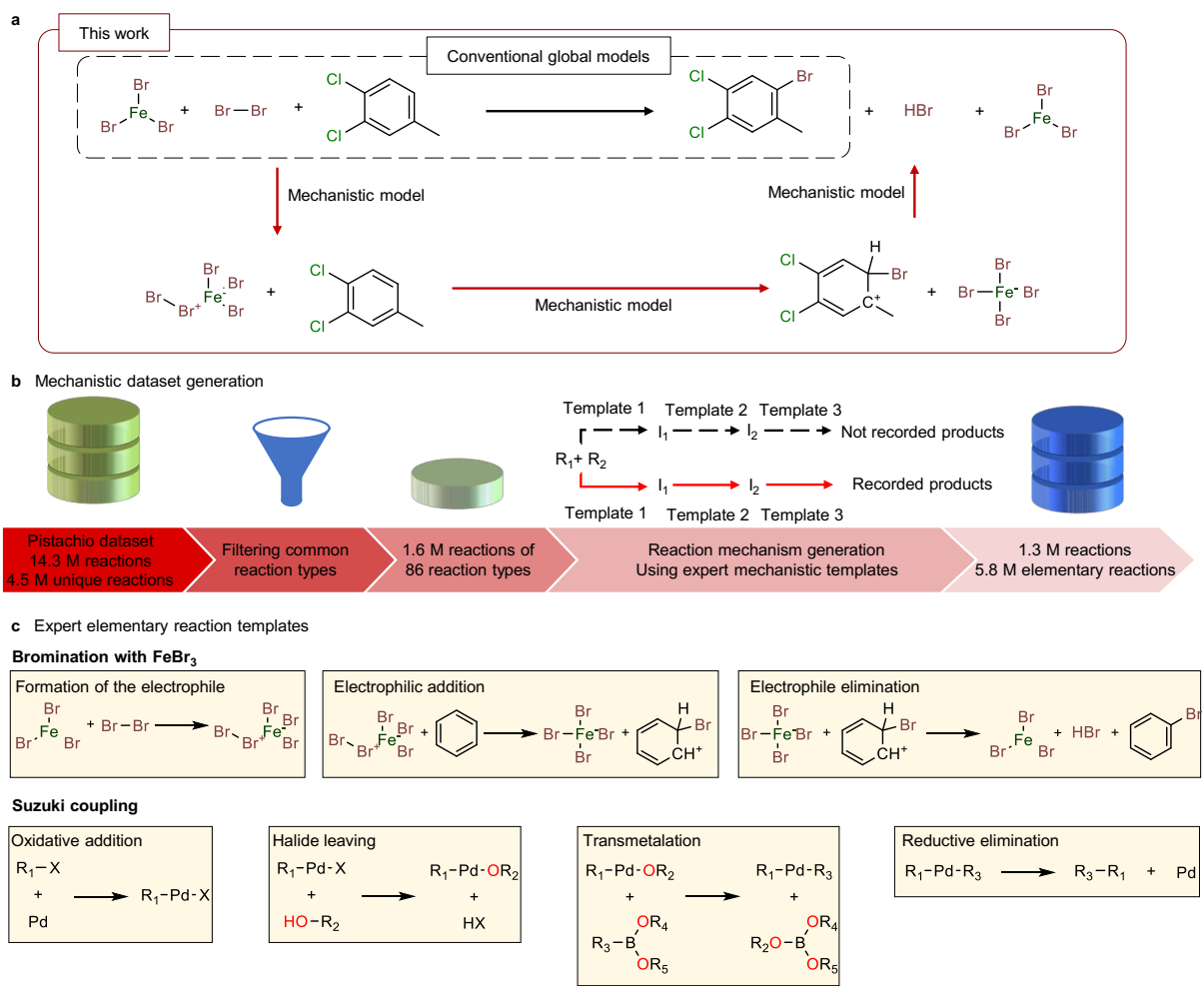

**Figure 1. a**. This work addresses the development of a mechanistic reaction prediction model, in contrast to prior work on reaction product prediction. **b**. Workflow for the generation of mechanistic dataset. Out of 4.5 million unique reactions in Pistachio, we focus on 1.6 million spanning the most popular 86 reaction classes. Expert elementary reaction templates were then applied to generate a mechanistic dataset of 5.8 million elementary reactions. **c**. Examples of expert elementary reaction templates. Additional reaction templates can be referenced in the Supplementary Information and all are available via Github.

**Results and discussion**

**Construction of mechanistic dataset with expert elementary reaction templates**

To build a new mechanistic dataset, we developed a set of elementary reaction templates that describe widely agreed upon mechanisms for the most popular reaction classes found in U.S. and European patents. For instance, a bromination reaction using $FeBr_3$ occurs through three elementary reactions: the formation of the electrophile, the electrophilic addition, and the electrophile elimination (Figure 1c). Elementary reaction templates are sequentially applied to experimentally-recorded reactants until the experimentally-recorded product is recovered, resulting in a mechanistic pathway containing imputed intermediates and byproducts. Due to the generality of the elementary reaction templates as written, they may lead to the generation of unexpected intermediates or side products, whether or not they are chemically plausible. Therefore, we prune the generated dataset to keep only the reaction pathways that lead to the recorded products (see the SI for the details).

We utilized the Pistachio 2022Q1 version to identify the 86 most popular reaction classes out of 1,756, covering approximately 30% of unique recorded reactions. As the same reaction class may have different reaction conditions such as different reagents, 175 different conditions were considered. This Pistachio dataset was subsequently divided randomly into training, validation, and test sets using an 8:1:1 ratio. Following this division, we applied our collection of elementary reaction templates to each reaction. We successfully established complete reaction mechanisms for 1,026,587, 145,996, and 145,710 reactions within the training, validation, and test sets respectively, corresponding to 4,482,889, 638,831, and 637,464 mechanistic steps. As this is a new dataset and new task, we focus on internal comparisons in our empirical evaluation.

**Machine learning models successfully recapitulate reaction mechanisms and final products**

We employed three distinct machine learning models: the Weisfeiler-Lehman difference network (WLDN), which formulates reaction prediction as the prediction of edits to a molecular graph;[4A] the Transformer, which formulates reaction prediction as the translation of reactant SMILES strings into product SMILES strings;[5A] and Graph2SMILES, which interprets reactant structures with a graph encoder but uses a sequence decoder to generate product SMILES strings.[6] Each model was trained to predict the next intermediate (or final product) in a mechanistic pathway. To identify when the reaction is considered complete, an additional elementary reaction template with no changes was appended at the end of each reaction sequence, explicitly indicating the end of the reaction. Therefore, the model predicts the termination of the reaction if it anticipates the same chemicals as the reactants.

Model performance is assessed based on the ability to accurately predict (a) the next structure at the elementary reaction level and (b) the entire sequence. Precision in predicting the correct next structure is crucial, but equally important is forecasting all intermediates in a multi-step sequence to correctly recover the final product. We measure the former through a top-k accuracy metric that measures the fraction of test cases where the true intermediate is predicted in the k highest-ranked predictions based on comparison of canonicalized SMILES strings. We measure the latter through a "sequence rank" metric, which measures the worst (i.e., highest) rank among the full sequence of elementary reactions in the pathway. The performance metrics for the full test set are summarized in Table 1.

.

**Table 1.** Performance on elementary reaction prediction for all three models on the test set. The full mechanistic data is randomly split into 8:1:1 training:validation:testing. The best performance for each column is bolded. Hyperparameters of the models are described in Section S2 and Table S1. For the sequence rank, the reaction pathways are pruned to avoid unproductive cycles.

|  | Top-k accuracy in elementary reaction (%) $N$=637,464 | | | | | Sequence rank (%) $N$=145,710 | | | | |
| --- | --- | --- | --- | --- | --- | --- | --- | --- | --- | --- |
|  | top-1 | top-2 | top-3 | top-5 | top-10 | top-1 | top-2 | top-3 | top-5 | top-10 |
| WLDN | 79.4 | 86.5 | 87.4 | 88.0 | 88.3 | 72.7 | 78.8 | 79.6 | 80.4 | 80.5 |
| Transformer | 83.5 | 88.3 | 89.3 | 90.1 | 90.7 | 76.1 | 87.1 | 88.6 | **89.9** | **90.6** |
| G2S | **88.8** | **91.0** | **91.3** | **91.5** | **91.6** | **83.7** | **88.5** | **88.8** | 89.0 | 89.1 |

The top-1 accuracy in elementary reaction prediction of 80-90% is comparable to the performance of product prediction models trained on similar patent-derived datasets.[4A, 5A, 6] However, the overall accuracy for sequence rank is lower than the top-k accuracy of elementary reactions, reflecting the impact of any inaccuracies in predicting individual elementary reactions on the overall sequence rank. The Graph2SMILES model achieves the best top-1 sequence rank, demonstrating its effectiveness in predicting the entire reaction sequence, but the Transformer achieves substantial improvements for top-2 through top-10 predictions.

**Model predictions of reaction mechanisms are inherently interpretable**
Conventional models predicting final products directly from reactants lack the ability to explain how products are formed. Moreover, these models can predict final products even without considering necessary catalysts or reagents for the reaction. In contrast, our models, trained with the mechanistic dataset, offer the capability to elucidate how products are formed, identify the involved reagents, and in principle determine favorable reaction conditions. This section demonstrates the explanatory power of our models by examining various examples of reaction pathways predicted by the WLDN model (Figure 2).

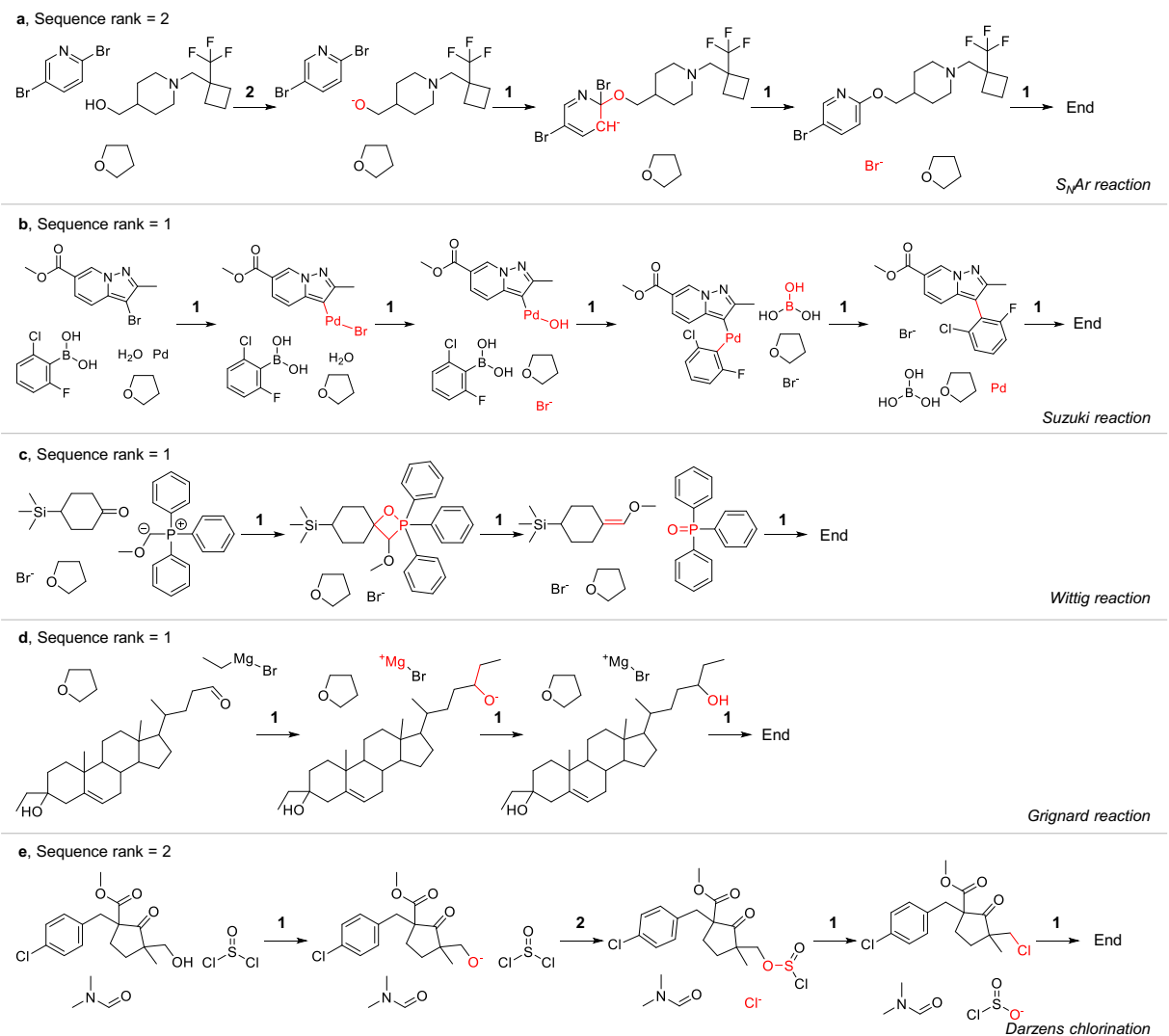

**Figure 2.** Correct predictions from the test set illustrating that the WLDN model has learned to generate multi-step reaction mechanisms. Atoms and bonds highlighted in red emphasize the chemical changes predicted at each step. The number above the arrow represents the rank of the elementary reaction predicted by the model.

**Suggesting the need to consider reaction conditions**: In Figure 2a, model predicts the preparation of a nucleophile for a nucleophilic aromatic substitution ($S_NAr$) reaction by deprotonating an alcohol as the first step. Based on how we have defined our elementary reaction templates, models are able to predict acid-base reactions without explicitly specifying a proton source/sink, so it does not explicitly consider the presence of a base for alcohol deprotonation. However, it suggests to a user that a base may be warranted to run this pathway. After deprotonation, the model correctly predicts the subsequent reactions which are the formation of a Meisenheimer complex and the loss of bromide lead to the final product.

**Understanding the role of reagents and regeneration of catalysts**: Figure 2b shows a Suzuki reaction, where a palladium catalyst is crucial. The model accurately identifies key steps of the Suzuki reaction, including oxidative addition, transmetalation, and reductive elimination. Notably, the model predicts the recovery of $Pd^0$ in the fourth elementary reaction. These catalysts are often neglected by models solely focused on predicting final products from reactants as their structure is unchanged in the final products.[4A, 7] Consequently, models designed to primarily capture major chemical changes may struggle to identify the involvement of catalysts. This limitation enables such models to predict final products even when catalysts are absent, which is not chemically plausible. This highlights the importance of our mechanistic approach in understanding the role of catalysts (See Section S3 and Table S2 for the details).

Reagents actively contribute to the chemical alterations like catalysts, but are transformed into byproducts. These reagents are crucial for the initiation of the reactions. The mechanistic model accurately predicts how these reagents react as shown in the Grignard reaction in Figure 2d and Darzens chlorination in Figure 2e and Figure S4. This predictive capability shows the model's performance in identifying the distinct roles of various chemical components.

**Generation of possible reaction byproducts**: The Wittig reaction, as in Figure 2c, produces an alkene from a ketone and an ylide. Here, the model predicts the key intermediate of oxaphosphetane and the final products. Importantly, the model identifies the byproduct triphenylphosphine oxide ($Ph_3PO$) formed after the second elementary reaction (as well as the boric acid formed in the Suzuki reaction in Figure 2b), which are often ignored by other models predicting Wittig and Suzuki reactions. Our approach dictates every elementary reaction and is designed to track every heavy (non-hydrogen) atom, therefore, it enables the model to track the destination of all chemicals in the reaction to anticipate the formation of byproducts.

**Mechanism prediction identifies possible impurities of the reaction**

The byproducts and side products of chemical reactions may include undesirable impurities that are important to identify for the design of subsequent isolation strategies or consideration of one-pot compatibility. Preliminary computational approaches have treated the prediction of minor products as simply the prediction of lower-ranked "major products".[4A, 18] However, these models do not account for byproducts and still lack explainability. Mechanistic models may offer a solution to this challenge by predicting the reaction mechanism leading to impurity formation (Figure 3).

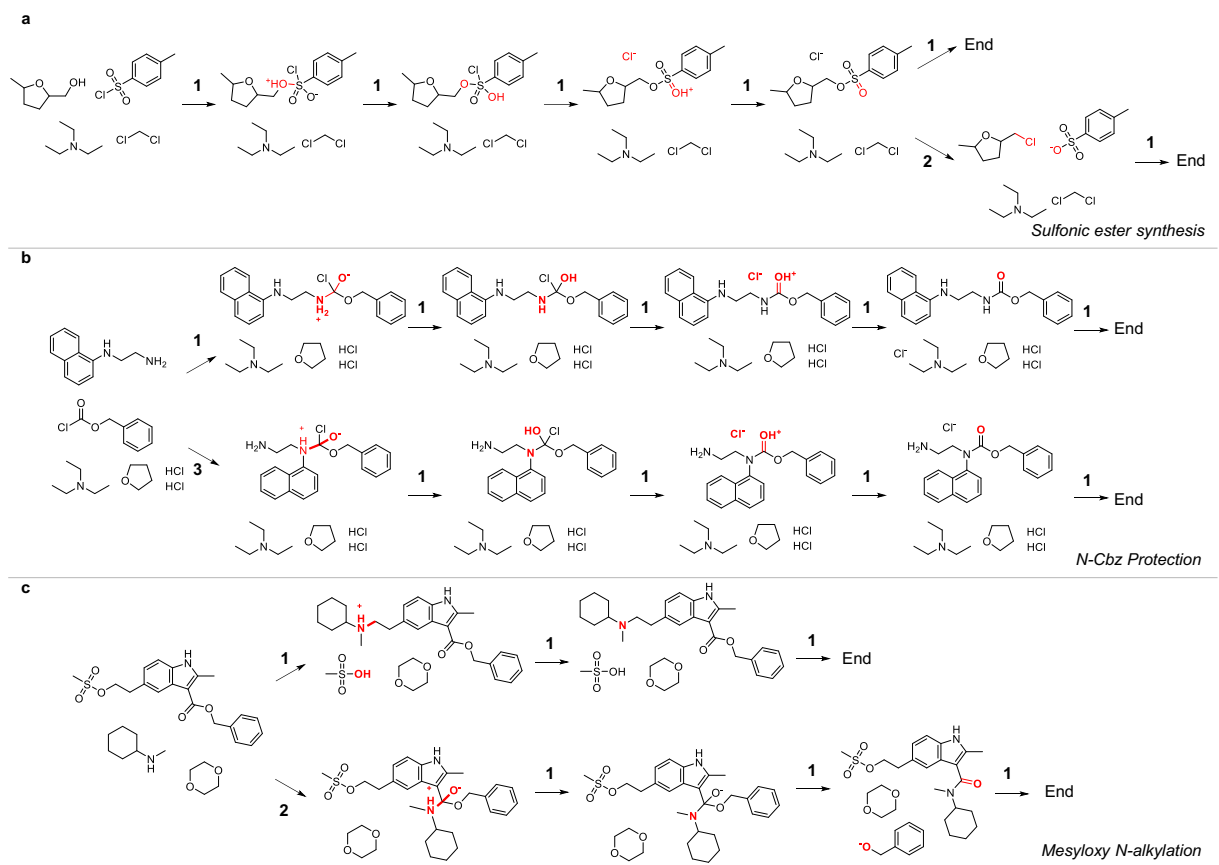

**Figure 3.** Impurity pathways predicted by the mechanistic WLDN model. Atoms and bonds in red emphasize the chemical changes according to the WLDN model prediction. The number above the arrow represents the rank of the reaction predicted by the model.

Sulfonate ester formation using sulfonyl chloride as a reactant generates chloride ions as a byproduct in the final reaction mixture (Figure 3a). Depending on the reaction conditions, there is a possibility that chloride could undergo further reactions with sulfonate esters, resulting in the formation of sulfonates and alkyl chlorides as impurities.[19]

In Figure 3b, the reactants containing two amines may follow two distinct pathways of N-Cbz protection using benzyl chloroformate. The mechanistic model identifies that the primary amine is more reactive than the secondary amine. While the reaction in Figure 3b illustrates a competing reaction within the same functional group, Figure 3c depicts the competition between two different reactions of mesyloxy N-alkylation and amide formation. The model successfully predicts the major product of mesyloxy N-alkylation, which is a reaction class not present in the training set.

Impurities are generally not reported in published chemical reactions, nor tabulated in reaction databases, causing challenges for accurate quantitative evaluations. However, our study demonstrates a proof of principle for anticipating impurities associated with specific reaction pathways by employing mechanistic models. This approach will enable a more comprehensive understanding of reaction outcomes, especially concerning the formation of unintended byproducts.

**Failure modes of mechanistic reaction prediction**

We investigate the common failures of mechanistic models in Figure 4. The Transformer and Graph2SMILES models, as sequence generative models, do not enforce atom conservation in the way that the WLDN model does when predicting graph edits. In Figure 4a, the Transformer model successfully predicts the first two steps of N-Cbz protection but unexpectedly deletes the atoms highlighted in magenta during the third step. The mechanistic Transformer model appears to recognize leaving the chloride ion but fails to reconstruct a benzyl carbamate moiety. Such language-based models may randomly distort intermediate structures, leading to diminished overall performance.

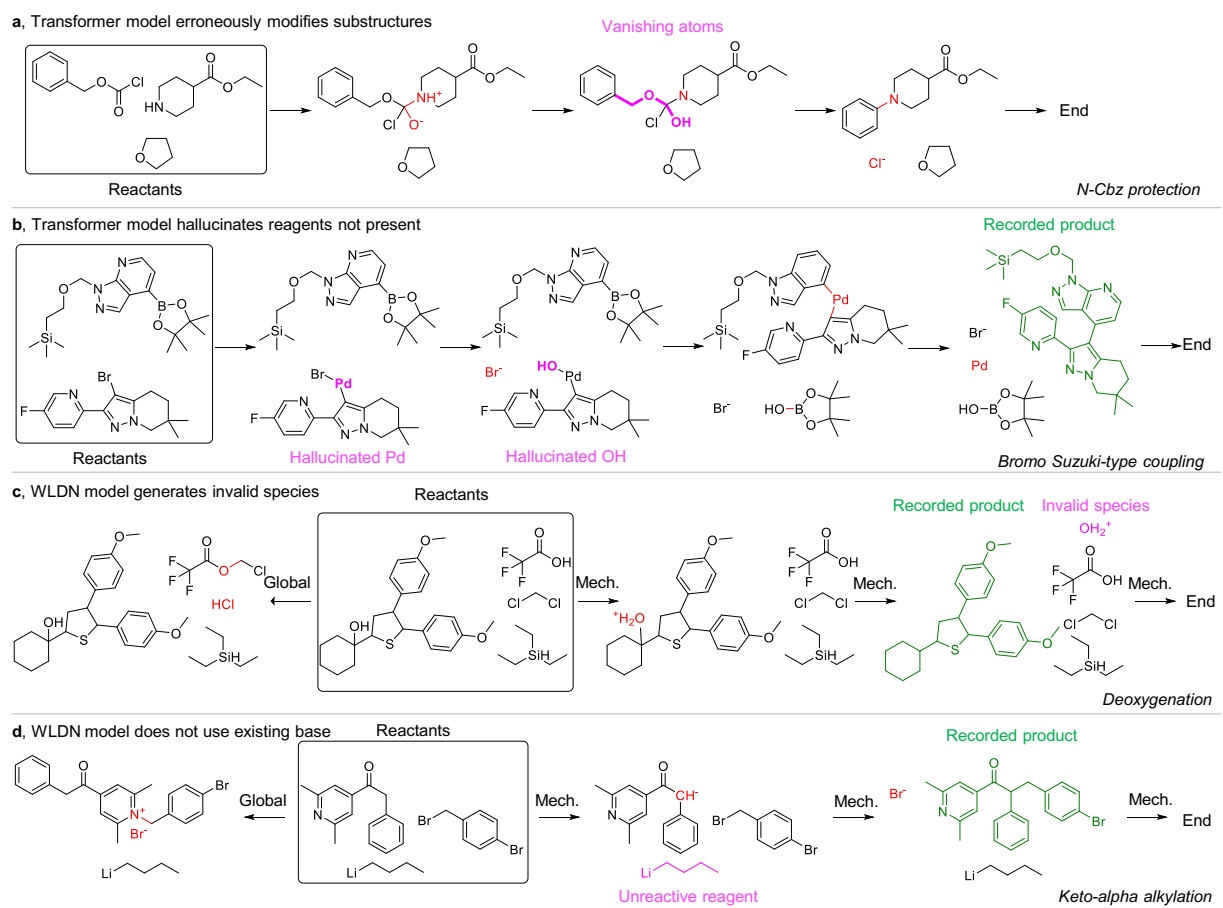

**Figure 4**. Demonstration of typical failure modes observed in mechanistic models. **a**. Failure in prediction by the Mechanistic Transformer model due to the deletion of atoms highlighted in magenta. **b**. Successful prediction by the mechanistic Transformer model of the recorded product while generating new atoms highlighted in magenta. **c**. Successful prediction by the mechanistic WLDN model of the recorded product through non-physical mechanisms, but global model fails to predict the product. **d**. Successful prediction by the mechanistic WLDN model of the recorded product while ignoring the organolithium base, but the global model fails to predict the product.

Violation of atom conservation can result in seemingly successful predictions, as illustrated in Figure 4b. "Suzuki-type couplings" in the Pistachio dataset refer to reactions resembling Suzuki reactions but lacking palladium catalysts. As indicated in Table S2, all global models predict the recorded outcomes of Suzuki-type couplings without palladium catalysts, despite this not being chemically reasonable. Conversely, the mechanistic WLDN model does not proceed with the reaction in the absence of palladium catalysts because the graph-based model is allowed to modify the bonds between the only existing atoms. In contrast, the mechanistic Transformer and Graph2SMILES models can predict the final products by hallucinating the presence of palladium to facilitate the oxidative addition step and generating a hydroxide ion for the transmetalation step (Figure 4b). This enables them to make a prediction that matches the recorded product at the expense of chemical validity.

While the mechanistic WLDN model does not create or delete atoms, it can still undergo non-physical and unrealistic reaction mechanisms. These discrepancies become apparent when applied to reaction types dissimilar to those it has been trained on, such as the deoxygenation illustrated in Figure 4c. The mechanistic WLDN model successfully predicts the recorded products but includes unrealistic mechanistic steps and invalid chemical species, such as a positively charged water molecule.

A similar pattern emerges in acid-base reactions, where the mechanistic dataset is compromised due to the complexity of accounting for all possible acid-base reactions. Therefore, the mechanistic models are taught to allow protonation/deprotonation steps without explicit acid/base sources. However, when suitable reagents are provided to the model, the mechanistic model may still ignore their role in the reaction, as exemplified in Figure 4d where butyl lithium is not used for alpha ketone deprotonation.

**Mechanistic models do not yet show the better generalizability to new reaction types**

As part of our original motivation for pursuing mechanistic models, we hypothesized that a model operating at the level of elementary reaction may have the potential to predict unseen reaction types through novel combinations of known elementary steps. Shared elementary steps could exist among different reaction types, even if the precise sequence of steps or the choice of functional groups is not shared. In this section, we evaluate the generalizability of the model by measuring the performance on unseen reaction classes. To directly compare the generalizability of mechanistic models to global models, we trained three additional machine learning models on the original reaction dataset prior to mechanistic imputation. While the global models attempt to predict the outcome in one step, mechanistic models need consecutive predictions to reach the final product. To ensure the successful recovery of the product, a beam search is employed for consecutive predictions in the mechanistic model (See the Section S4 for the details). The performances of all models on 14 holdout reaction classes not present in the training set are summarized in Figure 5.

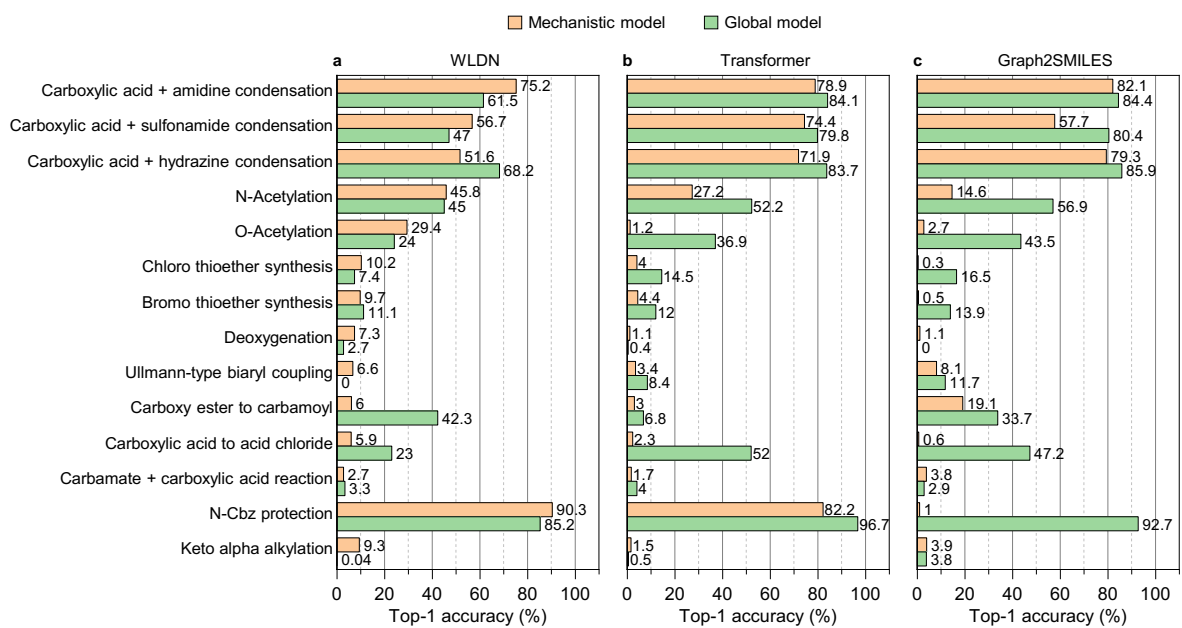

**Figure 5.** Comparison between top-1 accuracies of mechanistic and global models on 14 reaction classes not present in the training set.

The overall performance of mechanistic models on these unseen reaction classes is lower than that of global models. Mechanistic models demonstrate consistently better top-1 accuracy only for reactions involving carboxylic acids, consistent with the high frequency of reactions involving carboxylic acids in the training dataset. Despite our original hypothesis, there is no evidence that mechanistic reaction prediction models exhibit better generalization to new reaction types than traditional reaction prediction models.

This limitation can be attributed to several factors, including the potential missing reagents and stoichiometric details, the constrained diversity within the training set, challenges associated with consecutive predictions, and the violation of atom conservation in language models. Mechanistic models depend on detailed information from reactants, catalysts, reagents, and their stoichiometric details for accurate predictions, but the lack of detailed evaluation for missing reagents in holdout reactions can lead to poor performance. In contrast, global models trained on final products may better handle missing details. The models have been trained on training set with limited diversity, covering nearly 30% of reactions from the Pistachio dataset but based only on 86 common reaction types out of 1,767. This limitation in diversity restricts its ability to recognize or understand certain molecules not included in the training set, leading to a narrow scope of applicability. Consecutive prediction is challenging, as errors can accumulate across multiple predictions if the model encounters unfamiliar intermediates, leading to inaccuracies. Addressing these issues will be crucial for improving the model's performance and applicability.

**Conclusions**

In this work, we present a new mechanistic dataset generated using expert elementary reaction templates, and present machine learning models for mechanism prediction. Our investigation into mechanistic models for chemical reaction prediction reveals both their capabilities and challenges. Mechanistic models exhibit promising accuracy in predicting the details of elementary reactions through an inherently interpretable problem formulation. Their ability to predict reaction mechanisms provides valuable insights into the roles of catalysts, reagents, and the intricate steps involved. Our exploration of impurity predictions shows the potential of mechanistic models to elucidate the formation pathways of unintended byproducts and side products, addressing a critical aspect often overlooked by conventional global models. Furthermore, the capacity of mechanistic models to predict reactive chemical species throughout reactions opens the opportunity to identify and discover novel reactions, but currently only in principle.

The generalizability of mechanistic models compared to global models remains an ongoing challenge and warrants additional investigation. Challenges such as difficulties in consecutive predictions and atom conservation violations, especially evident in language models like Transformer and Graph2SMILES, demonstrate the need for further improvements. The future of mechanistic models lies in overcoming these challenges, curating a more diverse mechanistic dataset, enhancing consecutive prediction accuracy, and improving the ability to fully conserve mass and charge throughout the reaction sequences.

**Acknowledgements**

This work was supported by the Machine Learning for Pharmaceutical Discovery and Synthesis consortium and the National Science Foundation under Grant No. CHE-2144153. The authors acknowledge the MIT SuperCloud and Lincoln Laboratory Supercomputing Center that have contributed to the research results reported within this paper.

**Conflicts of interest**

The authors declare no conflict of interest.

**Data Availability Statement**

The codes for the mechanistic dataset generation and machine learning models are available on GitHub: https://github.com/jfjoung/mechanism_prediction.

Supporting Information for

# Beyond Major Product Prediction: Reproducing Reaction Mechanisms with Machine Learning Models Trained on a Large-Scale Mechanistic Dataset

Joonyoung F. Joung[1], Mun Hong Fong[1], Jihye Roh[1], Zhengkai Tu[2], John Bradshaw[1], and Connor W. Coley[1, 2, *]

[1]Department of Chemical Engineering, Massachusetts Institute of Technology, Cambridge, Massachusetts 02139, United States

[2]Department of Electrical Engineering and Computer Science, Massachusetts Institute of Technology, Cambridge, Massachusetts 02139, United States

*E-mail: ccoley@mit.edu (C.W.C.)

## S1. Mechanistic dataset

The construction of the mechanistic dataset involved the utilization of expert elementary reaction templates. For instance, the nucleophilic aromatic substitution is described using three reaction templates as shown in Figure S1a. In the first step, the alcohol undergoes deprotonation (R1 in Figure S1b). It should be noted that acid/base reactions are not explicitly described with acid/base sources due to the challenges associated with considering all possible combinations of acid/base reactions as templates. The reactions illustrated in Figure S1b involve two reactants, where only one molecule possesses an alcohol group that can be deprotonated according to the first elementary reaction template. This template yields a deprotonated alcohol. The specificity of the second elementary reaction template was compromised to enhance generalizability, resulting in the production of six possible intermediate structures. The last elementary reaction template describes halide leaving, yielding six different chemical species. Application of the three elementary reaction templates to the two reactants produces six distinct chemicals as potential products. However, Pistachio records only one product, so we selected and preserved the reaction network linking the reactants to the recorded products, highlighted by red arrows in Figure S1b. Although a chloride ion is generated along with the recorded product, it is not documented in Pistachio. Two reactions, R8 and R9, were saved with the recorded product and the chloride ion.

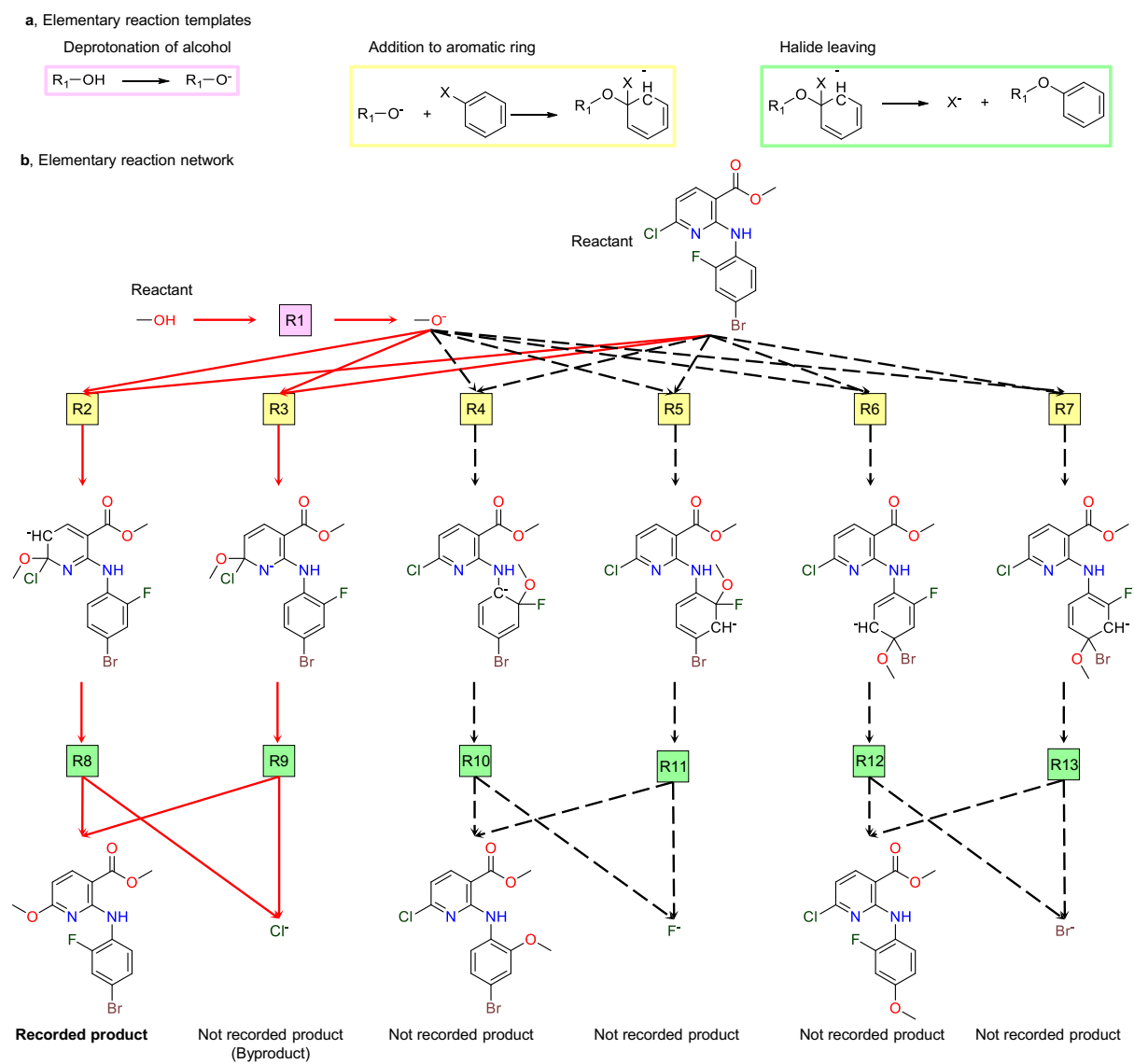

**Figure S1**. Illustration of elementary reaction templates and the network generated by their application to the reactants. **a**. Elementary reaction templates describing the nucleophilic aromatic substitution. **b**. Elementary reaction network constructed by application of the templates. All possible chemical structures produced by the templates are depicted.

We developed elementary reaction templates for the most popular 86 reaction classes as categorized in Pistachio. These 86 reaction classes out of 1,756 cover 30% of unique reactions. As the same reaction class may have different reaction conditions such as using different reagents, 175 different conditions were considered when we make the templates. For example, we examined two conditions for the condensation of carboxylic acids and amines: the reaction utilizing reagents like *N*,*N'*-dicyclohexylcarbodiimide as a catalyst and the reaction without any catalysts. Some reaction classes, such as chloro, bromo, and iodo N-alkylation reactions, may share the same elementary reaction templates.

On average, our approach achieved a product reproduction accuracy of approximately 69.6%. It's important to note that certain reactions in the dataset were deficient in crucial reagents, like water or hydroxide ions necessary for ester hydrolysis as shown in Figure S2a. Since our methods require all chemical species involved in reactions, products from reactions with missing essential reagents couldn't be reproduced using our mechanistic templates as depicted in Figure S2b.

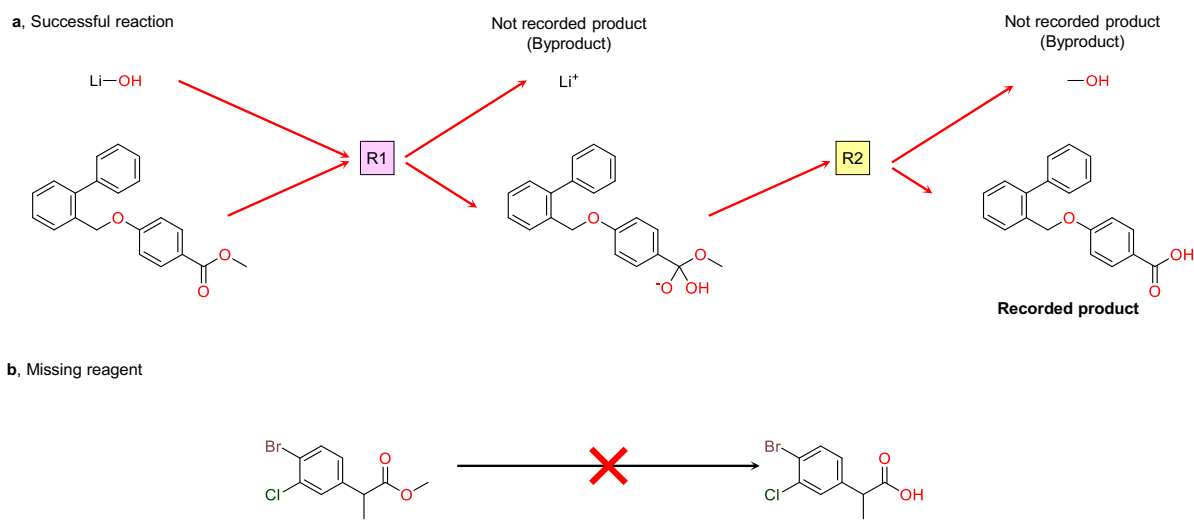

**Figure S2**. Significance of having all essential reagents in reactions. **a**. Successful reaction with all reactants and reagents. **b**. Reaction with a missing hydroxide source, leading to the inability to apply the elementary reaction templates.

## S2. Machine learning models

**WLDN model**: Our WLDN model adopts the primary structure from our earlier work.[1] This model divides the task into two steps: Core model as the first step determines which atoms are most likely to undergo changes. In the original model, only changes in connectivity were considered. However, intermediates formed during reactions may possess various formal charges or valences. To address this, we modified the first model to consider the number of hydrogen and formal charges. The model selects the top 16 changes with high likelihoods, preparing intermediate candidates by enumerating possible changes combinatorially. The final set of candidate intermediates is restricted to combinations of up to 6 unique changes. Connectivity constraints are also applied to reduce the number of valid candidates. In this process, 'no change' is also considered, explicitly indicating the end of the reaction. Candidate ranker, the second step, then determines the most likely molecular structure at the elementary reaction level. The hyperparameters of the model are summarized in Table S1.

**Transformer and Graph2SMILES models**: We used and adopted the models developed by Tetko and coworkers,[2] and our previous work.[3] Data augmentation, typically employed to increase model accuracy by generating equivalent random SMILES, was not employed in this study due to challenges in achieving identical production of intermediate structures. During augmentation using RDKit, certain properties such as charges, the number of hydrogen, or aromaticity may undergo sudden changes. While such changes may not pose a problem for neutral molecules like reactants or products, the intermediates could exhibit unusual valence, charge, or disrupted aromaticity. The hyperparameters of the models, taken from prior work without modification, are summarized in Table S1.

**Table S1**. Hyperparameters of the mechanistic models

**WLDN**

| | |
|---|---|
| Core – batch size | 10 |
| Core – Hidden size | 300 |
| Core – depth | 3 |
| Core – output dimension | 5 |
| Core – learning rate | 0.001 |
| Core – clipnorm | 5 |
| Candidate ranker – batch size | 1 |
| Candidate ranker – Hidden size | 500 |
| Candidate ranker – depth | 3 |
| Candidate ranker – learning rate | 0.001 |
| Candidate ranker – clipnorm | 5 |

**Transformer**

| | |
|---|---|
| Number of Transformer layers for both encoders and decoders | 4 |
| Hidden size | 512 |
| Attention heads in Transformer layers | 8 |
| Filter size in Transformer layers | 2048 |
| Batch size (by token count) | 4096 |
| Accumulation count | 4 |
| Train steps | 1250000 |
| Hidden dropout | 0.1 |
| Attention dropout | 0.1 |
| adam_beta1 | 0.9 |
| adam_beta2 | 0.998 |
| Learning rate factor | 2.0 |
| Learning rate schedule | Noam |
| Warmup steps | 8000 |

**Graph2SMILES**

| | |
|---|---|
| Embedding size | 256 |
| Hidden size | 256 |
| Filter size in Transformer | 2048 |
| Number of D-MPNN layers | 4 |
| Number of D-GAT attention heads | 8 |
| Attention encoder layers | 6 |
| Attention encoder heads | 8 |
| Decoder layers | 6 |
| Decoder heads | 8 |
| Number of accumulation steps | 8 |

**S3. Understanding the role and regeneration of catalysts**

As indicated in the main text, the global models exhibit the ability to predict final products even in the absence of catalysts, whereas the mechanistic models lack this capability. This distinction becomes evident in Suzuki-type couplings, specific reaction classes in NameRxn where a palladium catalyst is absent.[4] The performances of global and mechanistic models, particularly their top-1 accuracies in iodo-, bromo-, and chloro-Suzuki-type couplings, are summarized in Table S2. In general, global models demonstrate better performance on Suzuki-type couplings compared to mechanistic models. As shown in Figure S3, the global models predicts the products successfully without catalyst. However, the mechanistic WLDN model predicts C-Br dissociation instead of accurately forecasting intermediates of the traditional Suzuki coupling pathway. This is evident in its inability to predict the product of any Suzuki-type couplings, as summarized in Table S2, resulting from the constraint of atom conservation. The mechanistic Transformer and Graph2SMILES models may predict the final products by hallucinating the catalysts not present in reactants as described in the main text. We would argue that successful prediction of these "Suzuki-type" couplings in the absence of palladium is not necessarily a positive result.

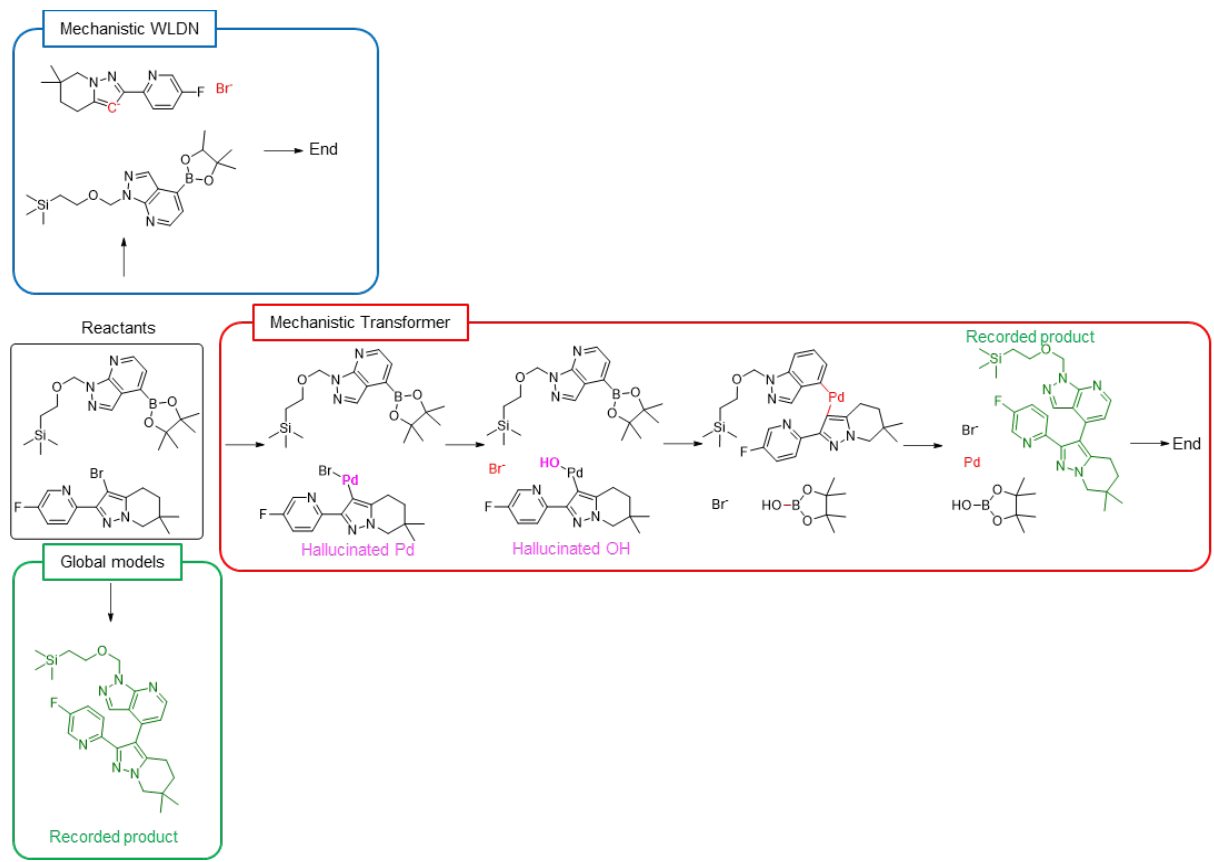

**Figure S3**. Predictions of the mechanistic and global models on Suzuki-type couplings.

**Table S2.** Comparison between top-1 accuracies of mechanistic and global models on Suzuki-type couplings.

| Reaction class name | WLDN | | Transformer | | Graph2SMILES | |
|---|---|---|---|---|---|---|
| | Mech. | Global | Mech. | Global | Mech. | Global |
| Iodo Suzuki-type coupling | 0.0 | **85.4** | 40.0 | 25.2 | 6.0 | 57.4 |
| Bromo Suzuki-type coupling | 0.0 | **82.5** | 15.1 | 24.6 | 0.0 | 62.9 |
| Chloro Suzuki-type coupling | 0.0 | **81.6** | 8.1 | 31.2 | 0.0 | 64.9 |

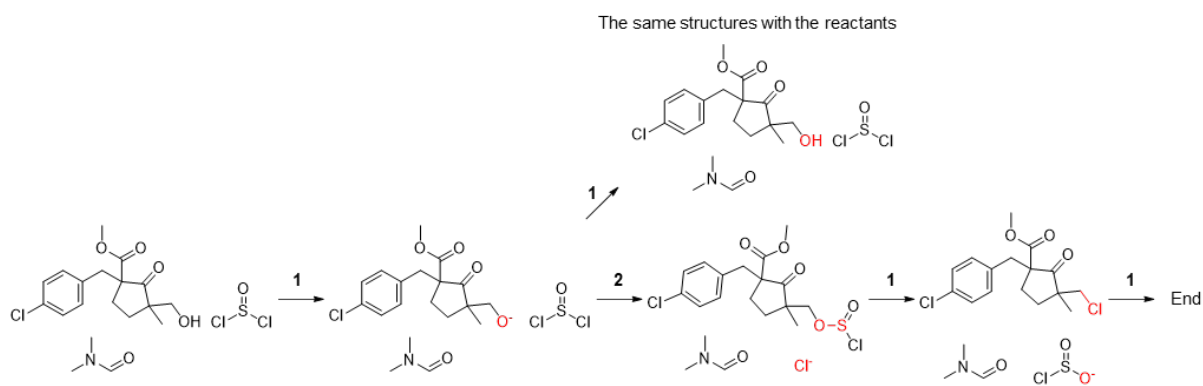

**Figure S4.** Predictions from the test set illustrating that the WLDN model has learned to generate multi-step reaction mechanisms. Atoms and bonds highlighted in red emphasize the chemical changes predicted at each step. The model assigned a higher probability to the prediction of the reaction involving protonation of the deprotonated site, thereby reverting to the reactant state, compared to the actual ground truth reaction.

## S4. Beam search

The WLDN predicts each intermediate step with an associated probability, which we use to guide a beam search based on the multiplication of probabilities towards the final nodes. In contrast, the transformer and Graph2SMILES models do not provide probabilities for candidates, but ranks can be obtained. For these models, the beam search is based on the rank, and to differentiate through depth, we introduced a discounting factor ($\gamma = 0.5$) to calculate the accumulated rank $R = \sum_n \gamma^n r_n$ where $r_n$ is the rank and $n$ is the depth. If the model predicts the same chemical species as the reactants, it serves as a stop sign in the prediction process.

During the beam search, we prune the reaction pathways if its reverse reaction forms a set of previously observed reactants.

## S5. Global models

We trained global models with overall reactions to compare mechanistic model. The performances on test set reactions are summarized in Table S3. The overall reaction dataset shares the same reactions as the mechanistic dataset but lacks the intermediates. Top-k accuracies of all three models are high because the reaction classes used for training are limited or less diverse.

**Table S3**. Performances (%) in reaction prediction of global models trained on overall reactions.

|  | top-1 | top-2 | top-3 | top-5 |
|---|---|---|---|---|
| WLDN | 95.0 | 97.1 | 97.5 | 97.6 |
| Transformer | 90.2 | 93.8 | 94.8 | 95.5 |
| G2S | **98.3** | **98.6** | **98.7** | **98.7** |

# Table of contents of elementary reaction templates

# Reaction class name: Carboxylic acid + amine condensation

**I. Condensation using DCC**

Required agent: 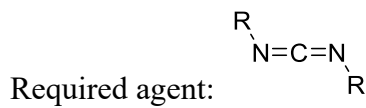

Elementary reaction

1. Proton exchange

   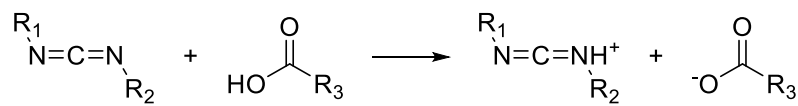

2. Formation of a single bond between carboxylic acid and protonated DCC

   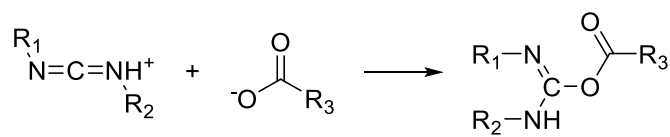

3. Addition of amine (alcohol) into carboxylic acid-DCC complex

   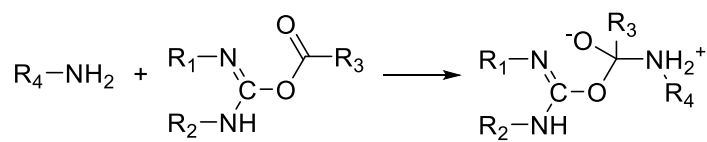

   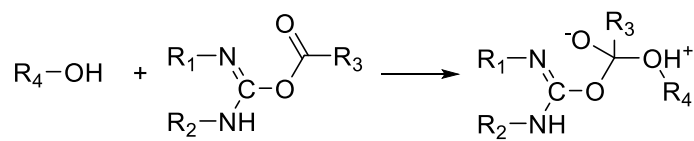

4. Cleavage into amide and urea

   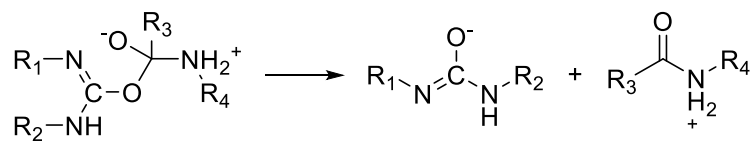

   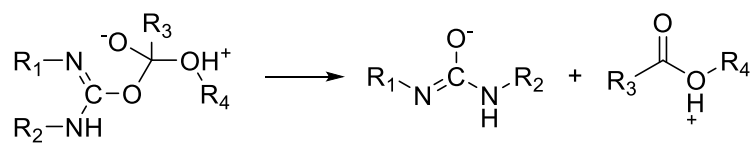

5. Proton exchange between amide and urea

   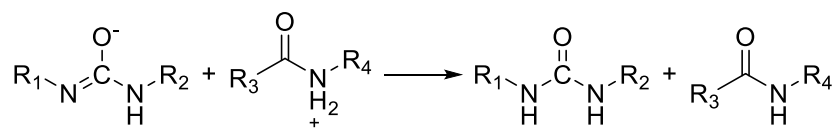

## II. Condensation without catalyst

Required agent: None

Elementary reaction

1. Addition of amine into carboxylic acid

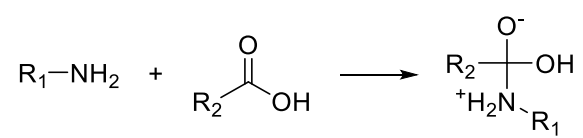

2. Deprotonation of amine

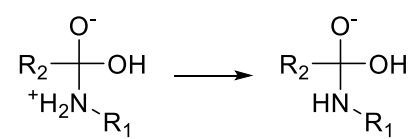

3. Hydroxide ion leaves

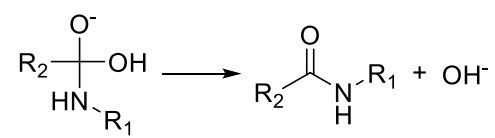

# Reaction class name: N-Boc deprotection

## I. Deprotection with OH⁻

Required agent: OH⁻

Elementary reaction

1. Addition of hydroxide ion

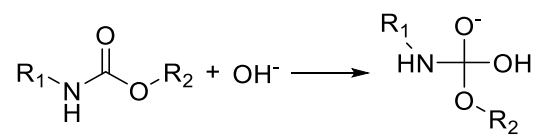

2. Alkoxide leaves

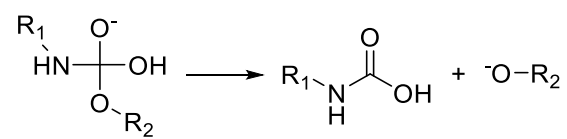

3. Deprotonation of carboxylic acid

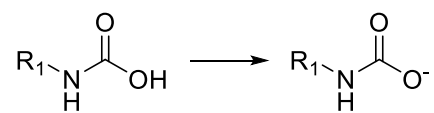

4. Evolution of CO2 and production of amide anion

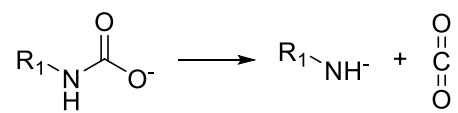

5. Protonation of amide anion

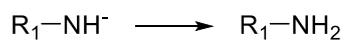

## II. Deprotection with Alkali hydroxide

Required agent: NaOH

Elementary reaction

1. Addition of hydroxide ion

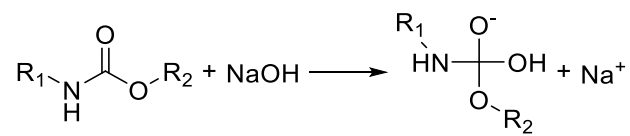

2. Alkoxide leaves

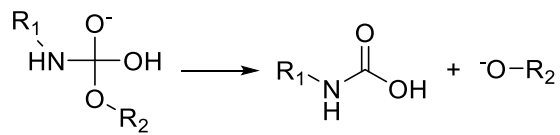

3. Deprotonation of carboxylic acid

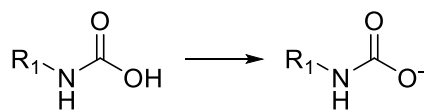

4. Evolution of CO2 and production of amide anion

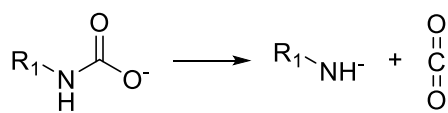

5. Protonation of amide anion

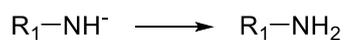

## III. Deprotection with water

Required agent: $H_2O$

Elementary reaction

1. Protonation of carbonyl group

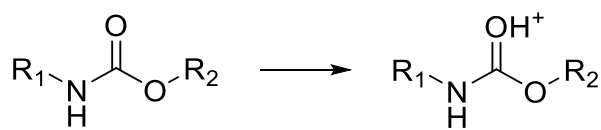

2. Addition of water into carbonyl group

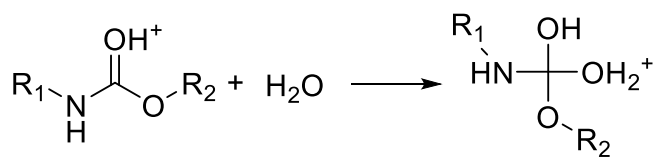

3. Deprotonation of protonated alcohol group

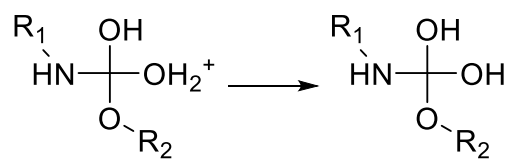

4. Alkoxyl group leaves

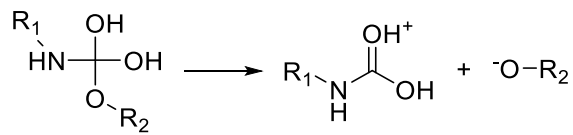

5. Proton exchange

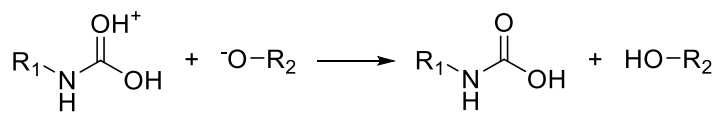

6. Deprotonation of carboxylic acid

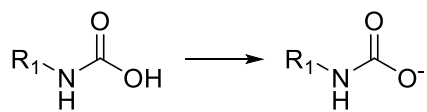

7. Evolution of CO2 and production of amide anion

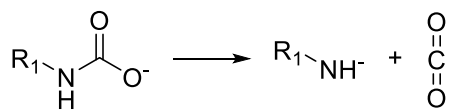

8. Protonation of amide anion

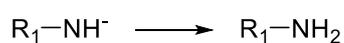

## IV. Non aqueous solution (Acidic condition)

Required agent: $H_2O$

Elementary reaction

1. Protonation of ester or carbonyl oxygen

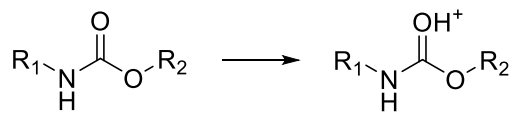

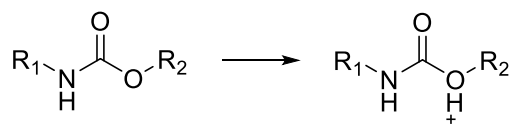

2. Carbocation leaves

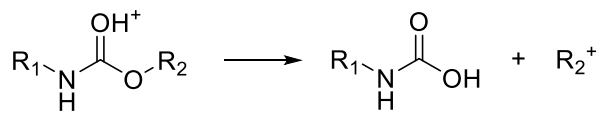

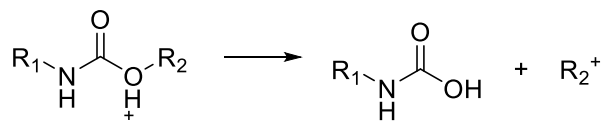

3. HOCO leaves

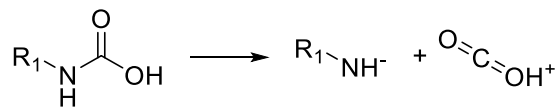

4. Proton exchange

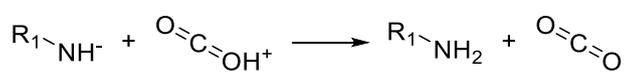

5. E1 or SN1 type reaction

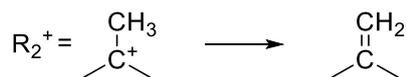

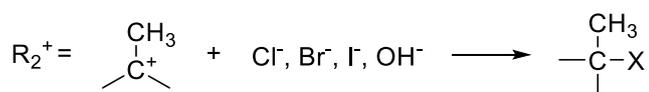

# Reaction class name: Williamson ether synthesis

## I. Reaction

Required agent: None

Elementary reaction

1. Addition of alcohol under acidic / deprotonation of alcohol

    Acidic condition

    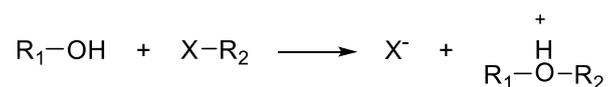

    Basic condition

    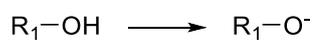

2. Neutralization of protonated ester / Addition of alcohol under basic conditions

    Acidic condition

    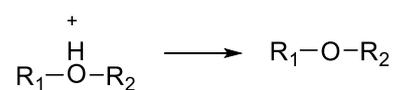

    Basic condition

    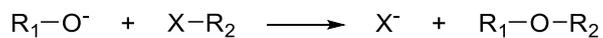

# Reaction class name: Bromo N-alkylation / Chloro N-alkylation / Iodo N-alkylation

## I. Reaction

Required agent: None

Elementary reaction

1. Addition of amine

   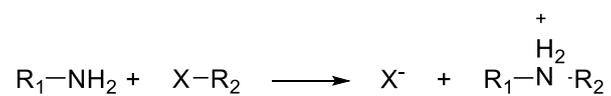

2. Amine deprotonation

   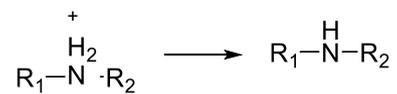

# Reaction class name: SNAr ether synthesis

## I. reaction with alcohol group

Required agent: None

Elementary reaction

1. Deprotonation of alcohol

   R₁–OH ⟶ R₁–O⁻

2. Addition to aromatic ring

   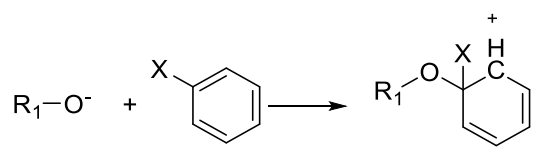

3. Leaving halide ion

   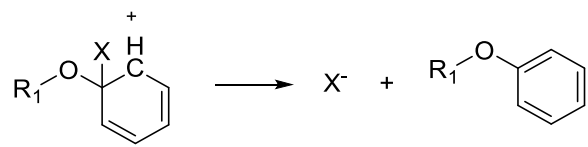

# Reaction class name: Aldehyde reductive amination

## I. Reaction

Required agent: None

Elementary reaction

1. Protonation of aldehyde

    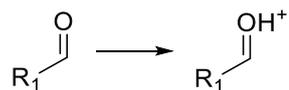

2. Addition of amine

    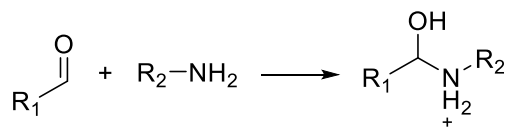

3. Proton exchange within the molecule

    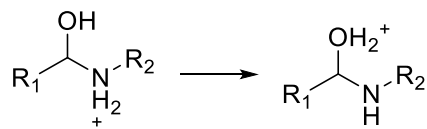

4. Water leaves

    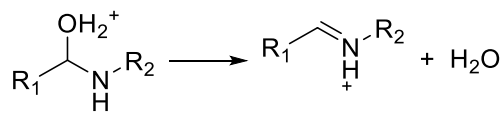

5. Reduction by borohydride or alumanuide

    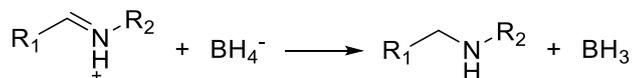

# Reaction class name: Bromination

## I. Bromination on benzene

Required agent: 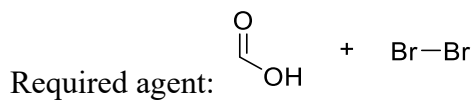  + Br—Br

Elementary reaction

1. Complexation of bromine and acetic acid

   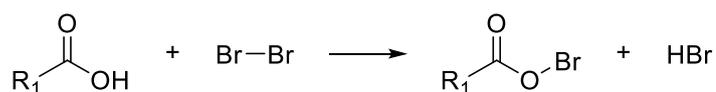

2. Bromination of aromatic ring

   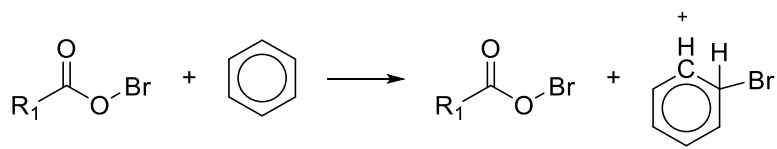

3. Deprotonation of aromatic ring

   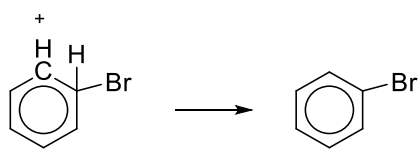

## II. Bromination on alpha ketone

Required agent: 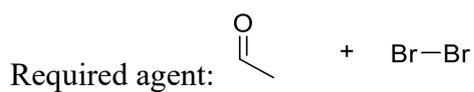  + Br—Br

Elementary reaction

1. Protonation of carbonyl

   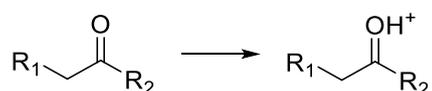

2. Deprotonation of alpha position carbon

   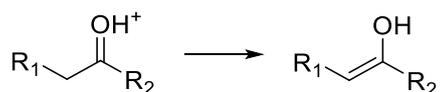

3. Bromination of alpha carbonyl

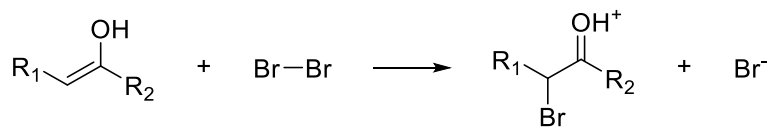

4. Deprotonation of carbonyl

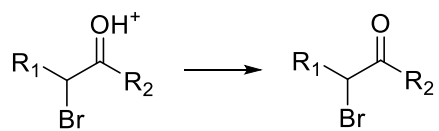

### III. Using FeBr₃ or AlBr₃

Required agent: 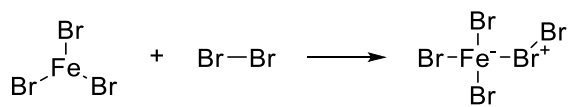

Elementary reaction

1. Activation of Lewis acid

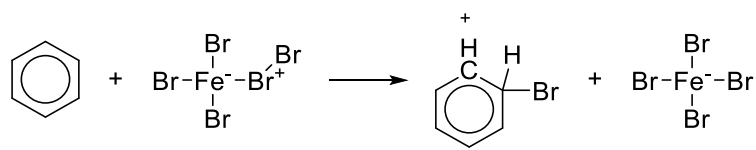

2. Bromination of aromatic ring

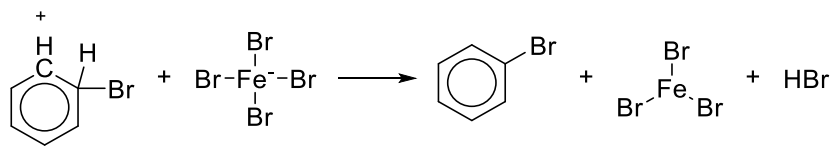

3. Proton exchange

### IV. Using NBS

Required agent:

Elementary reaction

1. Halogenation of aromatic ring

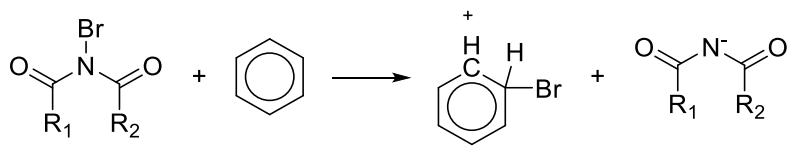

2. Proton exchange

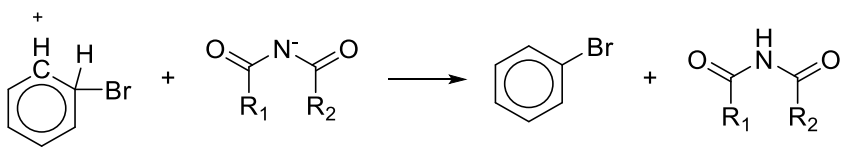

# Reaction class name: N-Boc protection

## I. Condensation using DCC

Required agent: None

Elementary reaction

1. Amine reacted with dicarbonate

   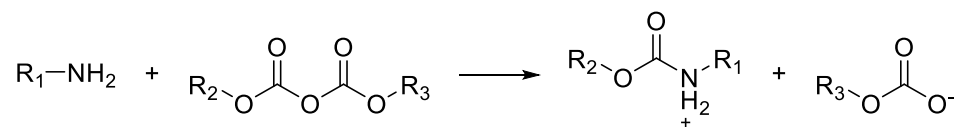

2. Carbon dioxide released

   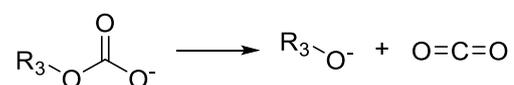

3. Proton exchange

   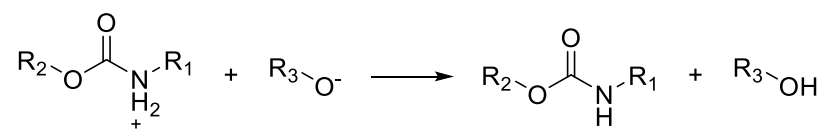

# Reaction class name: Carboxylic ester + amine reaction

## I. Reaction

Required agent: None

Elementary reaction

1. Addition of amine into carbonyl

   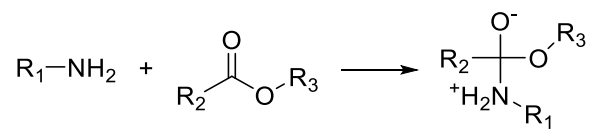

2. Deprotonation of amine

   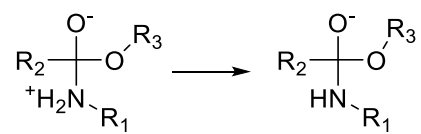

3. Alkoxide ion leaves

   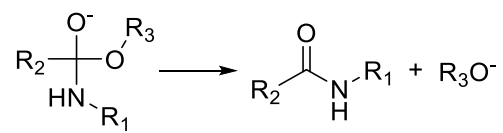

# Reaction class name: Carboxy to carbamoyl

## I. Condensation using DCC

Required agent: 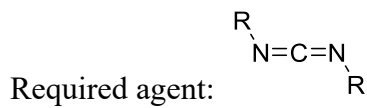

Elementary reaction

1. Deprotonation of amine

   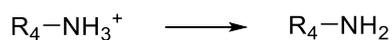

2. Proton exchange

   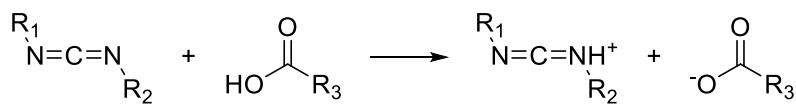

3. Formation of a single bond between carboxylic acid and protonated DCC

   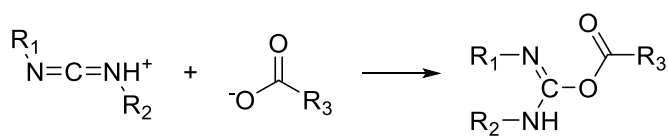

4. Addition of amine (alcohol) into carboxylic acid-DCC complex

   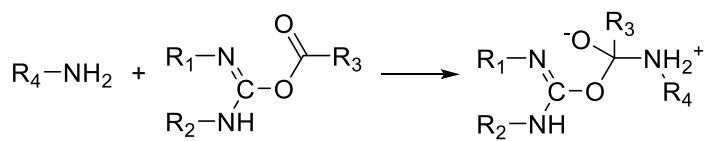

   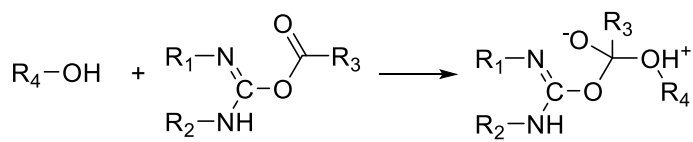

5. Cleavage into amide and urea

   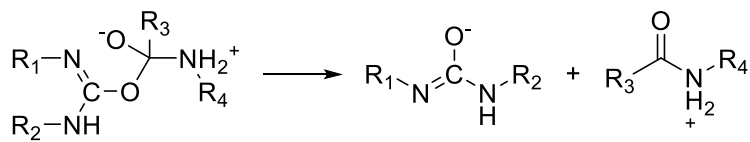

   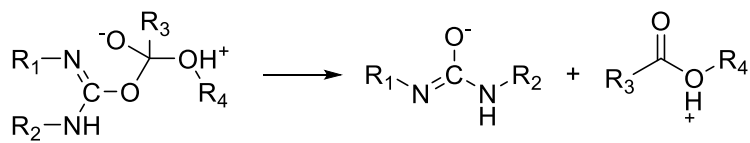

6. Proton exchange between amide and urea

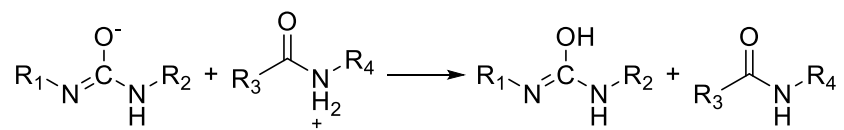

## II. Condensation without catalyst

Required agent: None

Elementary reaction

1. Deprotonation of amine

   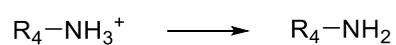

2. Addition of amine into carboxylic acid

   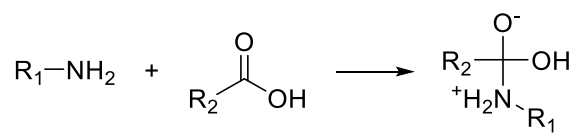

3. Deprotonation of amine

   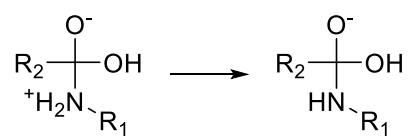

4. Hydroxide ion leaves

   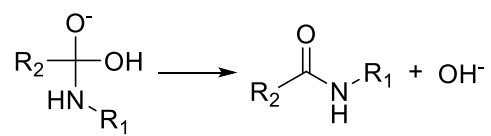

# Reaction class name: Esterification / Steglich esterification / Fischer-Speier esterification

## I. Condensation using DCC

Required agent: 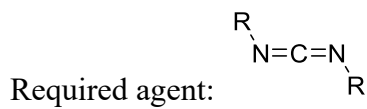

Elementary reaction

1. Proton exchange

   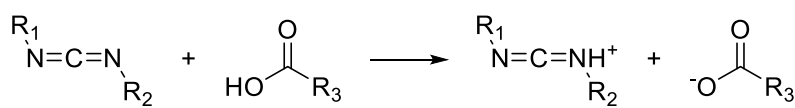

2. Formation of a single bond between carboxylic acid and protonated DCC

   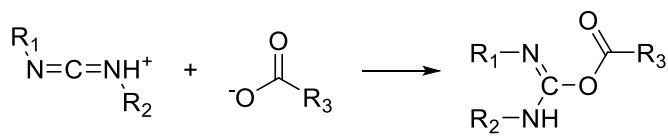

3. Addition of amine (alcohol) into carboxylic acid-DCC complex

   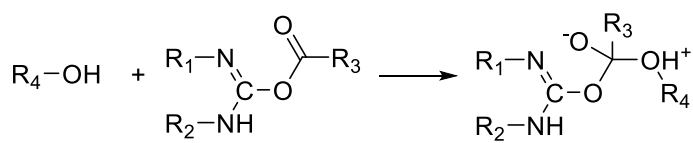

4. Cleavage into ester and urea

   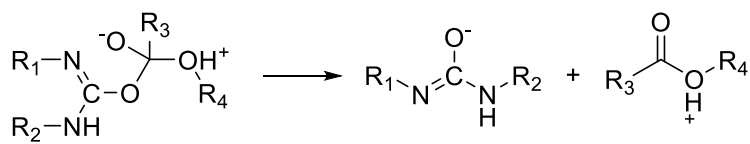

5. Proton exchange between ester and urea

   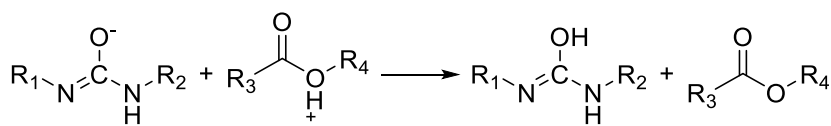

## II. Esterification

Required agent: 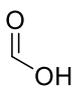

Elementary reaction

1. Protonation of carbonyl or deprotonation of alcohol

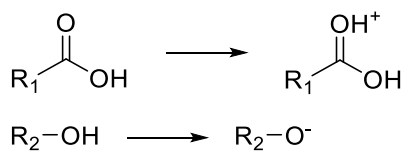

2. Alcohol addition to carbonyl

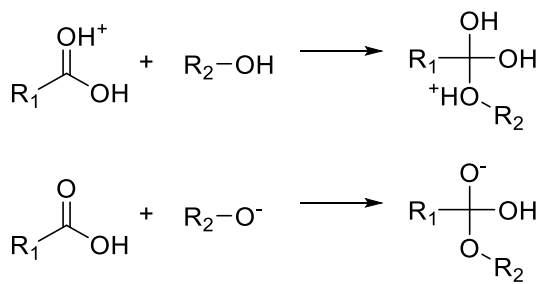

3. Protonation or deprotonation of complex

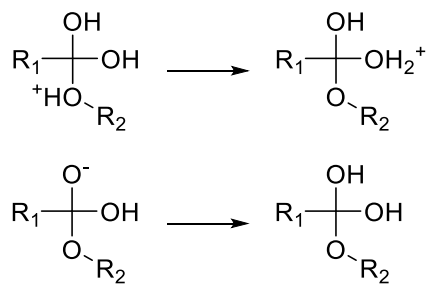

4. Water or hydroxide ion leaving

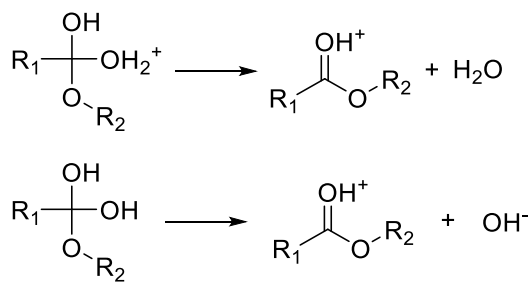

5. Proton exchange

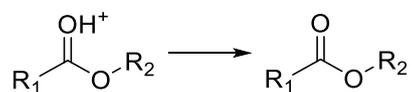

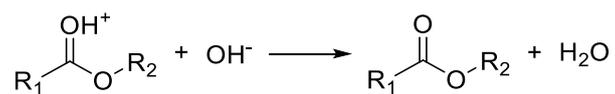 + OH⁻ → 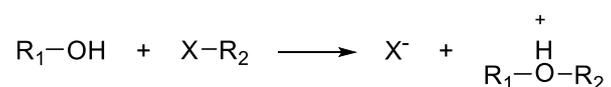 + H₂O

### III. Williamson ether synthesis type

Required agent: None

Elementary reaction

1. Addition of alcohol under acidic condition / deprotonation of alcohol

    Acidic condition

    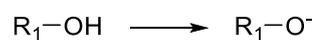

    $R_1-OH + X-R_2 \longrightarrow X^- + \overset{+}{\underset{R_1-O-R_2}{H}}$

    Basic condition

    $R_1-OH \longrightarrow R_1-O^-$

2. Neutralization of protonated ester / Addition of alcohol under basic conditions

    Acidic condition

    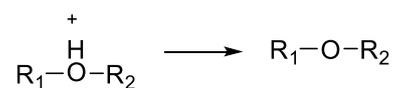

    $\overset{+}{\underset{R_1-O-R_2}{H}} \longrightarrow R_1-O-R_2$

    Basic condition

    $R_1-O^- + X-R_2 \longrightarrow X^- + R_1-O-R_2$

# Reaction class name: Methoxy to hydroxy

## I. Demethylation using BBr3

Required agent: 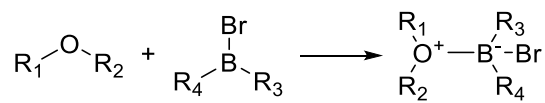

Elementary reaction

1. BBr3 addition

   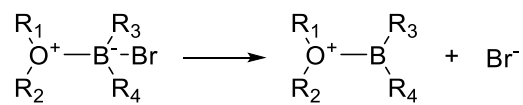

2. Halide leaving

   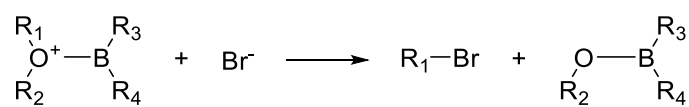

3. SN2 type alkyl halide leaving

   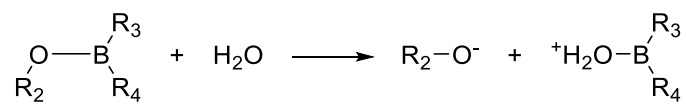

4. Addition of water

   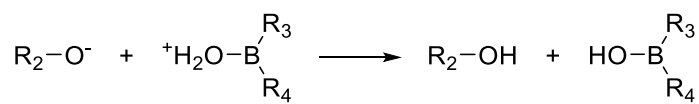

5. Proton exchange

   $R_2-O^- + {}^+H_2O-B\begin{matrix}R_3\\R_4\end{matrix} \longrightarrow R_2-OH + HO-B\begin{matrix}R_3\\R_4\end{matrix}$

## II. Ether acidic cleavage

Required agent: HBr

Elementary reaction

1. Proton exchange between ether and hydrohalic acid

   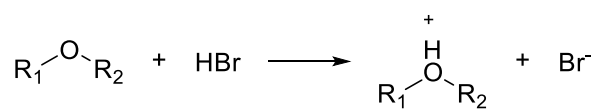

2. Cleavage of ether

R$_1$ and R$_2$ are primary or secondary carbon

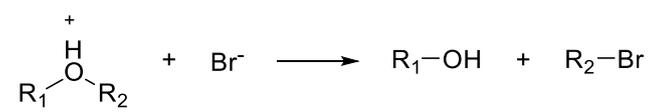

R$_1$ is tertiary carbon

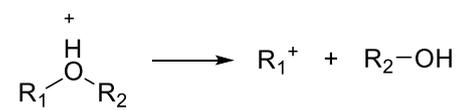

3. S$_N$1 type addition of halide

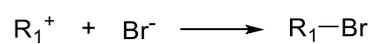

# Reaction class name: Ester to alcohol reduction

## I. Ester reduction using hydride

Required agent: 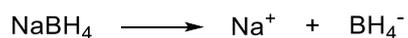

Elementary reaction

1. Preparation of Borohydride or Alumanuide

    NaBH$_4$ ⟶ Na$^+$ + BH$_4^-$

2. Hydride transfer

    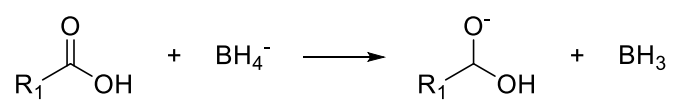

3. hydroxide or alkoxide leaving

    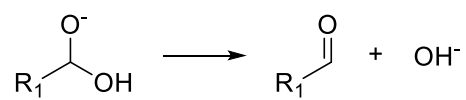

4. Additional hydride transfer

    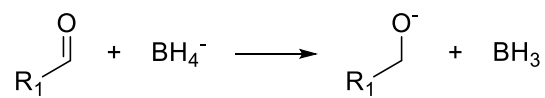

5. Protonation of alcohol

    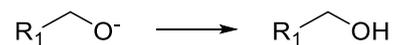

# Reaction class name: Bromo Suzuki coupling / Chloro Suzuki coupling / Iodo Suzuki coupling / Bromo Buchwald-Hartwig amination / Chloro Buchwald-Hartwig amination / Iodo Buchwald-Hartwig amination / Bromo Miyaura boration / Chloro Miyaura boration / Triflyloxy Miyaura boration / Iodo Miyaura boration

**I. Reaction with element Pd**

Required agent: Pd

Elementary reaction

1. Oxidative addition

    R$_1$—X  +  Pd  ⟶  R$_1$-Pd-X

2. Halide leaving

    R$_1$-Pd-X  +  HO—R$_2$  ⟶  R$_1$-Pd-OR$_2$  +  HX

3. Transmetalation

    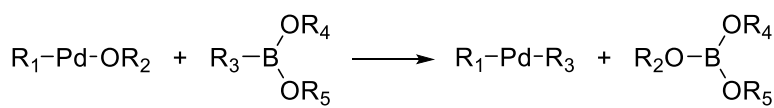

    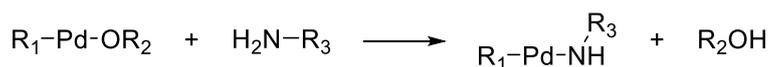

    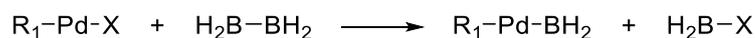

4. Reductive elimination

    R$_1$-Pd-R$_3$  ⟶  R$_3$—R$_1$  +  Pd

**II. Reaction with PdCl2**

Required agent: PdCl$_2$ and P

Elementary reaction

1. Palladium-ligand formation

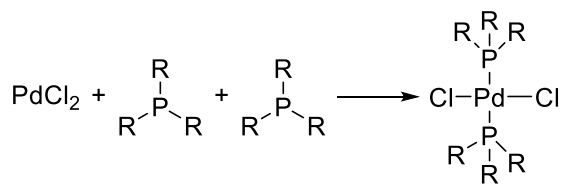

2. Palladium reduction

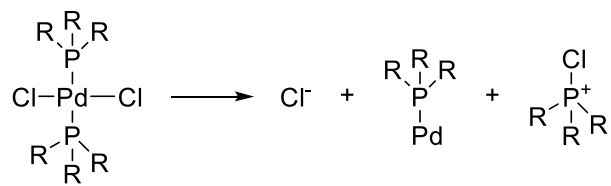

3. Oxidative addition

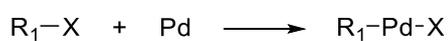

$R_1-X \ + \ Pd \ \longrightarrow \ R_1\text{-Pd-}X$

4. Halide leaving

$R_1\text{-Pd-}X \ + \ HO-R_2 \ \longrightarrow \ R_1\text{-Pd-}OR_2 \ + \ HX$

5. Transmetalation

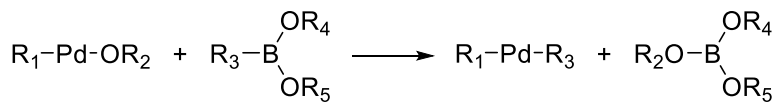

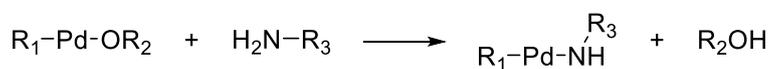

$R_1\text{-Pd-}X \ + \ H_2B-BH_2 \ \longrightarrow \ R_1\text{-Pd-}BH_2 \ + \ H_2B-X$

6. Reductive elimination

$R_1\text{-Pd-}R_3 \ \longrightarrow \ R_3-R_1 \ + \ Pd$

## III. Reaction with Pd coordinated with 3 or 4 ligands

Required agent: 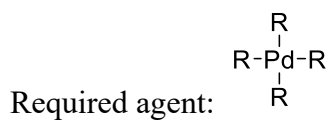

Elementary reaction

1. Oxidative addition

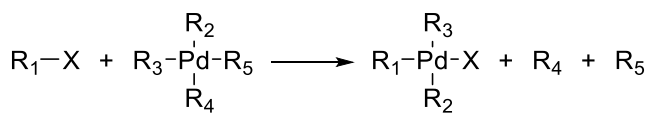

2. Halide leaving

   $R_1-Pd-X + HO-R_2 \longrightarrow R_1-Pd-OR_2 + HX$

3. Transmetalation

   $R_1-Pd-OR_2 + R_3-B(OR_4)(OR_5) \longrightarrow R_1-Pd-R_3 + R_2O-B(OR_4)(OR_5)$

   $R_1-Pd-OR_2 + H_2N-R_3 \longrightarrow R_1-Pd-NH(R_3) + R_2OH$

   $R_1-Pd-X + H_2B-BH_2 \longrightarrow R_1-Pd-BH_2 + H_2B-X$

4. Reductive elimination

   $R_1-Pd-R_3 \longrightarrow R_3-R_1 + Pd$

## IV. Reaction with Pd coordinated with 1 or 2 ligands

Required agent: $R_3-Pd-R_2$

Elementary reaction

1. Oxidative addition

   $R_1-X + R_3-Pd-R_2 \longrightarrow R_1-Pd(R_3)(R_2)-X$

2. Halide leaving

   $R_1-Pd-X + HO-R_2 \longrightarrow R_1-Pd-OR_2 + HX$

3. Transmetalation

   $R_1-Pd-OR_2 + R_3-B(OR_4)(OR_5) \longrightarrow R_1-Pd-R_3 + R_2O-B(OR_4)(OR_5)$

   $R_1-Pd-OR_2 + H_2N-R_3 \longrightarrow R_1-Pd-NH(R_3) + R_2OH$

   $R_1-Pd-X + H_2B-BH_2 \longrightarrow R_1-Pd-BH_2 + H_2B-X$

4. Reductive elimination

   $R_1-Pd-R_3 \longrightarrow R_3-R_1 + Pd$

# Reaction class name: Mitsunobu aryl ether synthesis / Mitsunobu imide reaction / Mitsunobu sulfonamide reaction / Mitsunobu ester synthesis

## I. Reaction

Required agent: None

Elementary reaction

1. Reaction between phosphine and diethylazodicarboxylate (DEAD)

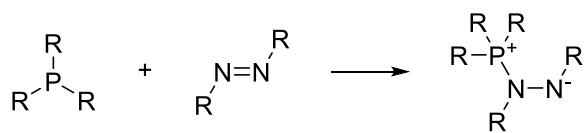

2. Proton exchange ($R_1OH$ p$K_a$ < 13)

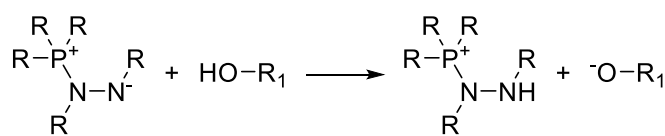

3. Forming a complex

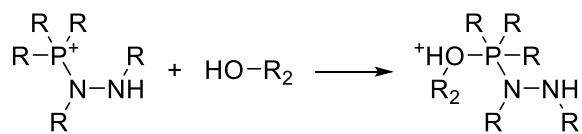

4. Bond cleavage and proton exchange

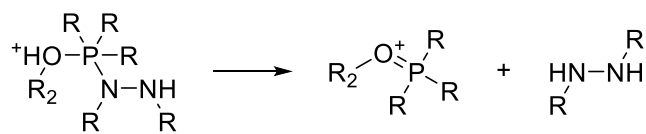

5. SN2 type reaction

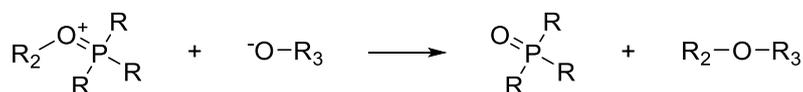

# Reaction class name: Ketone to alcohol reduction

## I. Reaction

Required agent: None

Elementary reaction

1. Hydride transfer

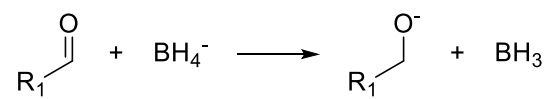

2. Protonation of alcohol

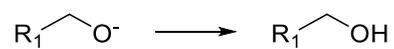

# Reaction class name: O-Bn deprotection

### I. Pd with H2

Required agent: Pd and H−H

Elementary reaction

1. Adsorption of ether on Pd

   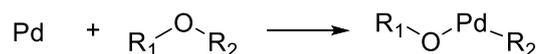

2. Hydrogenation

   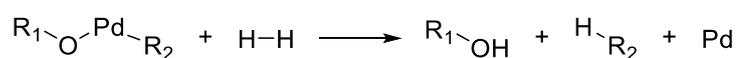

3.

### II. Deprotection using BBr3

Required agent: Br−B

Elementary reaction

1. BBr3 addition

   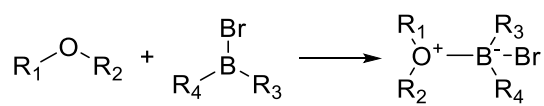

2. Halide leaving

   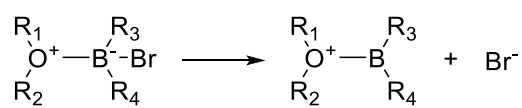

3. SN2 type alkyl halide leaving

   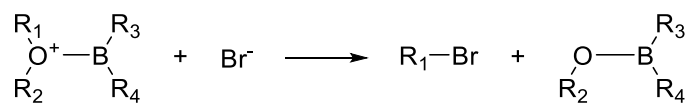

4. Addition of water

   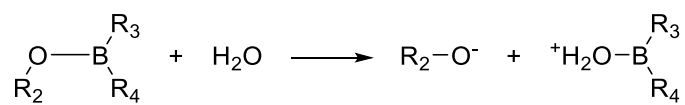

5. Proton exchange

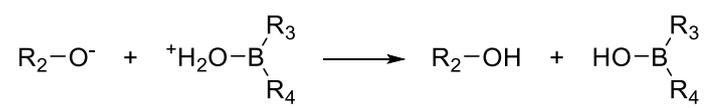

# Reaction class name: Isocyanate + amine urea coupling / Isocyanate + alcohol reaction

**I. Reaction**

Required agent: None

Elementary reaction

1. Amine addition to isocyanate

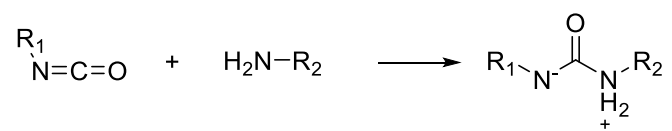

2. Proton exchange

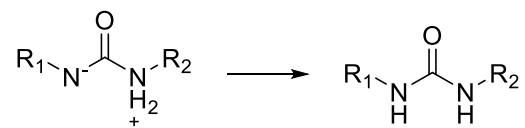

# Reaction class name: Ketone reductive amination

I. Reaction

Required agent: None

Elementary reaction

1. Amine addition

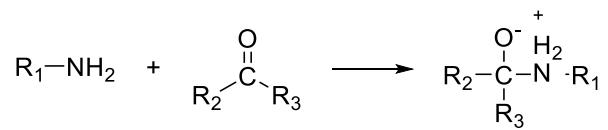

2. Proton exchange

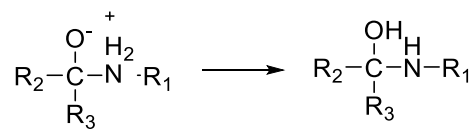

3. Protonation of alcohol

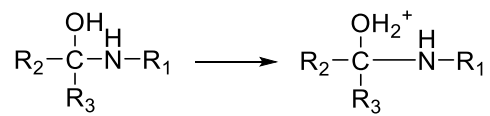

4. Water leaves

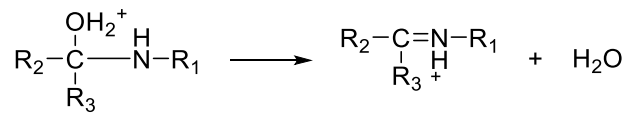

5. Hydride transfer

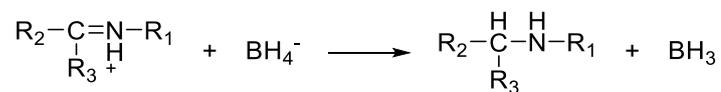

# Reaction class name: Sulfanyl to sulfonyl

## I. Reaction using peroxide

Required agent: 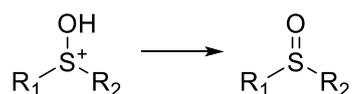

Elementary reaction

1. Oxidation

   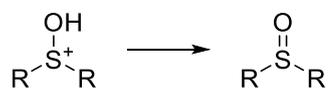

2. Deprotonation of S-OH

   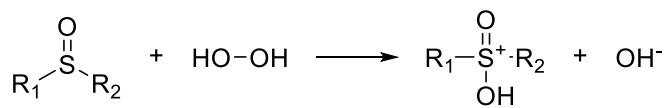

3. Additional oxidation

   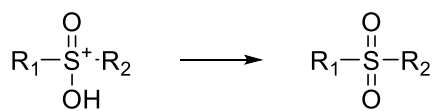

4. Deprotonation of S-OH

   

# Reaction class name: N-Cbz deprotection

**I. Strong acid**

Required agent: HBr

Elementary reaction

1. Proton exchange

   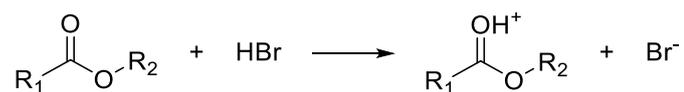

2. Nucleophilic substitution

   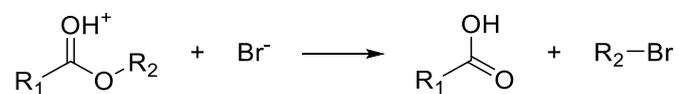

3. Deprotonation of carboxylic acid

   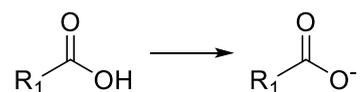

4. CO2 evolution

   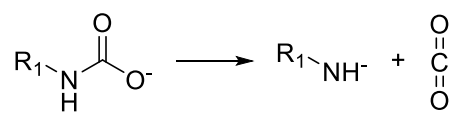

5. Protonation of amide anion

   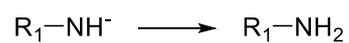

# Reaction class name: Iodo N-methylation

## I. Reaction

Required agent: None

Elementary reaction

1. Nucleophilic substitution

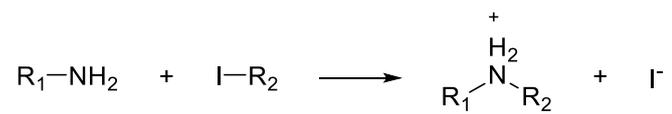

2. Deprotonation of amine

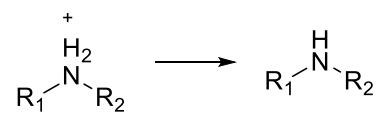

# Reaction class name: O-TBS deprotection / O-TBDPS deprotection / O-TIPS deprotection

## I. Fluoride condition

Required agent: HF

Elementary reaction

1. Fluoride addition

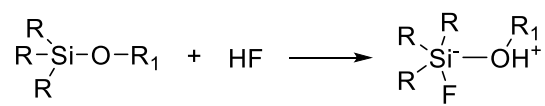

2. Si group leaves

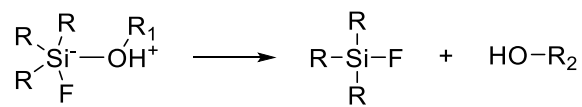

3. Protonation of alkoxide

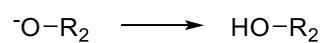

## II. Aqueous acidic condition

Required agent: $H_2O$

Elementary reaction

1. Protonation of ether

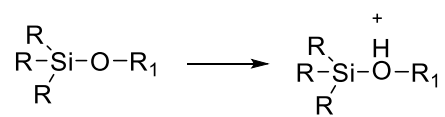

2. Nucleophilic substitution

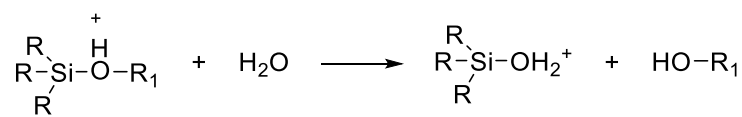

3. Deprotonation of protonated alcohol

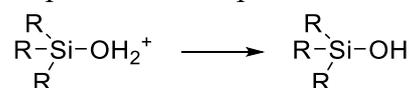

# Reaction class name: Methyl esterification

## I. Methyl iodide

Required agent: 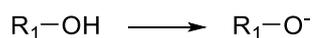

Elementary reaction

1. Deprotonation of carboxylic acid

    $R_1-OH \longrightarrow R_1-O^-$

2. Nucleophilic substitution

    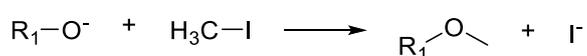

## I. Methanol

Required agent: $H_3C-OH$

Elementary reaction

1. Protonation of carbonyl or deprotonation of alcohol

    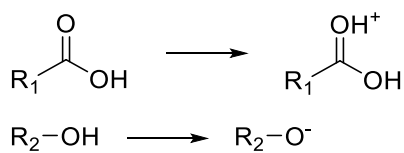

2. Alcohol addition to carbonyl

    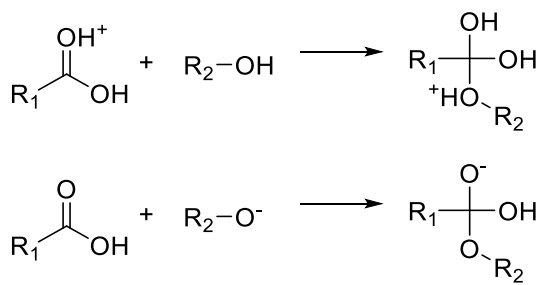

3. Protonation or deprotonation of complex

    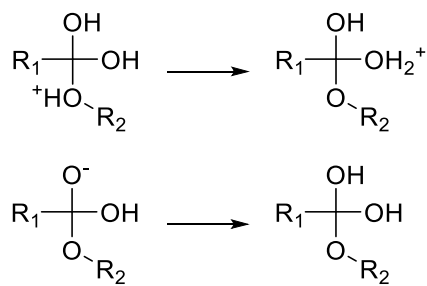

4. Water or hydroxide ion leaving

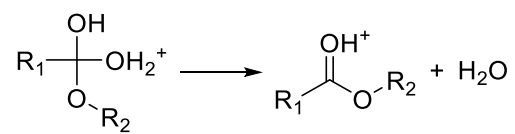

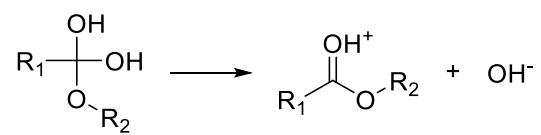

5. Proton exchange

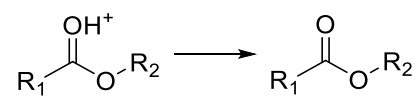

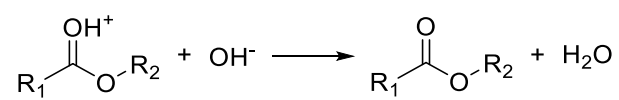

# Reaction class name: Chlorination

I. Using NCS

Required agent: 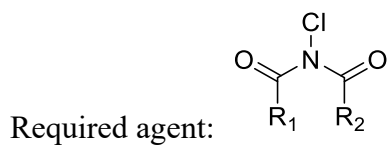

Elementary reaction

1. Halogenation of aromatic ring

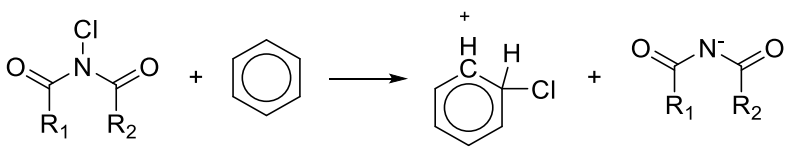

2. Proton exchange

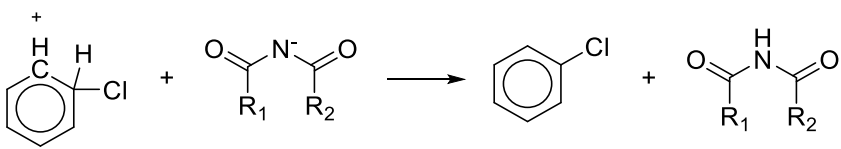

II. Using FeCl₃ or AlCl₃

Required agent: 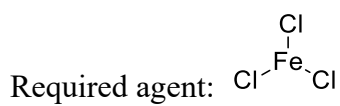

Elementary reaction

1. Activation of Lewis acid

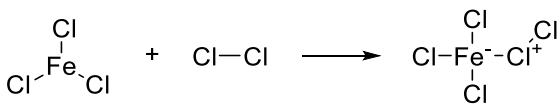

2. Bromination of aromatic ring

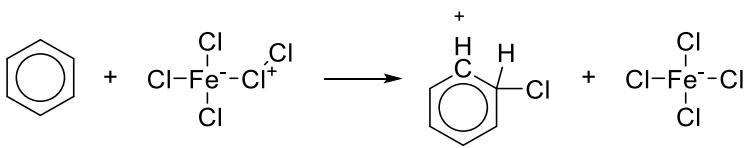

3. Proton exchange

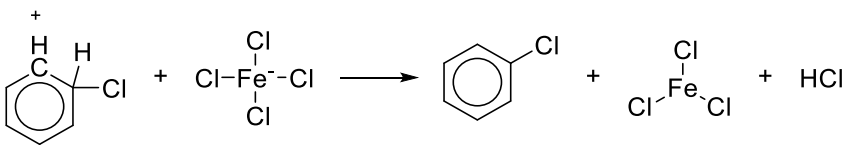

# Reaction class name: Hydroxy to chloro

**I. SOCl2 or POCl3**

Required agent: 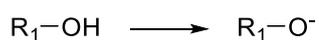

Elementary reaction

1. Deprotonation of alcohol

   $R_1-OH \longrightarrow R_1-O^-$

2. Addition of alkoxide

   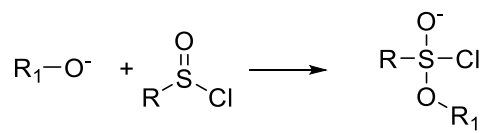

3. Chloride leaves

   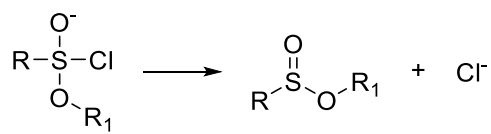

4. Nucleophilic substitution

   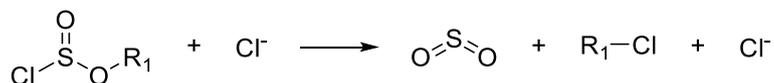

**II. PCl5**

Required agent: 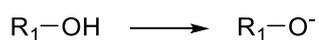

Elementary reaction

1. Deprotonation of alcohol

   $R_1-OH \longrightarrow R_1-O^-$

2. Addition of alkoxide

   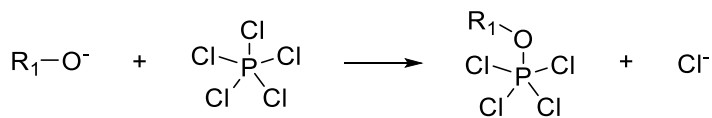

3. 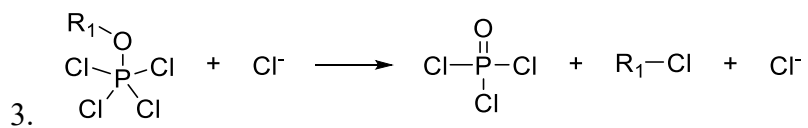

## III. Oxalyl chloride

Required agent: 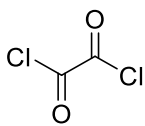

Elementary reaction

1. Deprotonation of alcohol

    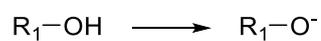

2. Addition of alkoxide

    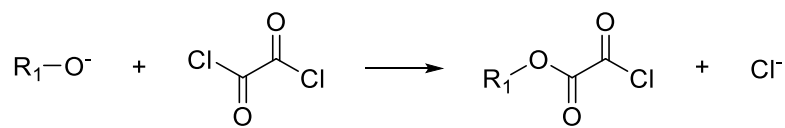

3. Nucleophilic substitution

    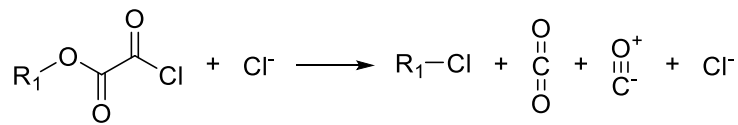

# Reaction class name: Iodination

## I. Using NIS

Required agent: 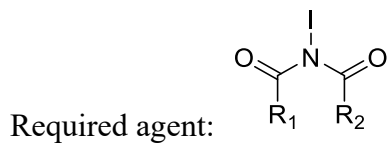

Elementary reaction

1. Halogenation of aromatic ring

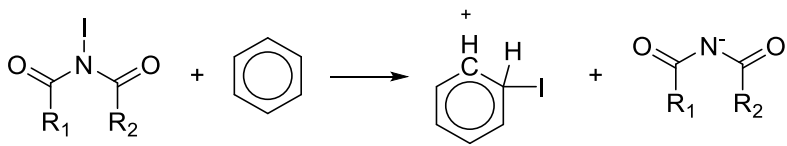

2. Proton exchange

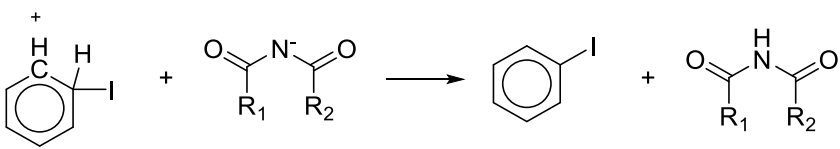

## II. Peroxide

Required agent: 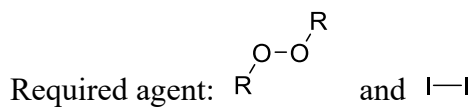 and I—I

Elementary reaction

1. Reaction of peroxide and iodine

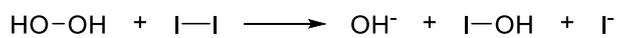

2. Halogenation of aromatic ring

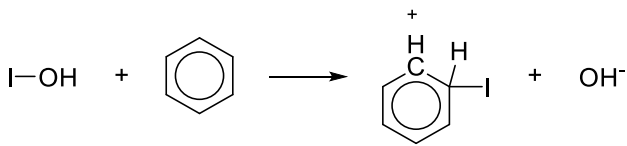

3. Proton exchange

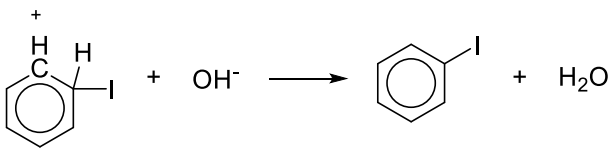

# Reaction class name: Amide to amine reduction

## I. Reaction

Required agent: $BH_4^-$

Elementary reaction

1. Preparation of Borohydride or Alumanuide

    $NaBH_4 \longrightarrow Na^+ + BH_4^-$

2. Hydride transfer

    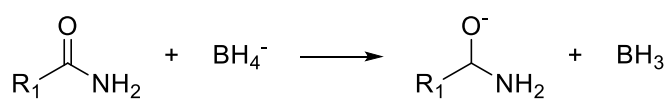

3. Salt formation

    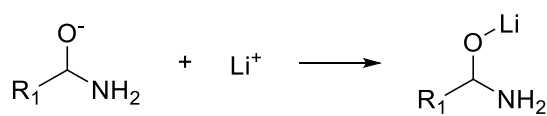

4. Boronic acid formation

    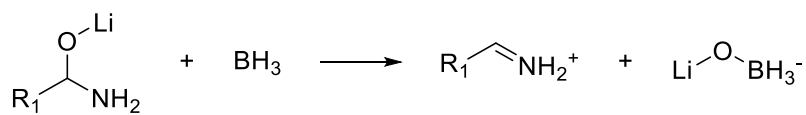

5. Additional hydride transfer

    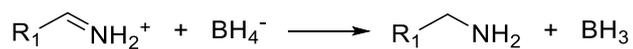

# Reaction class name: Alcohol + amine condensation

## I. Mitsunobu style reaction

Required agent: 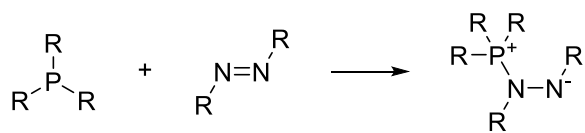

Elementary reaction

1. Reaction between phosphine and diethylazodicarboxylate (DEAD)

   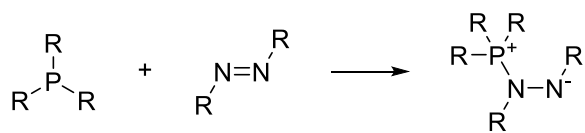

2. Proton exchange ($R_1OH$ p$K_a$ < 13)

   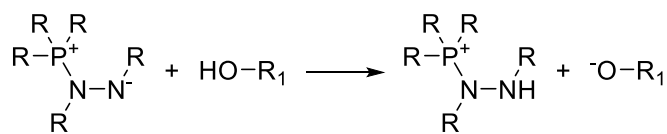

3. Forming a complex

   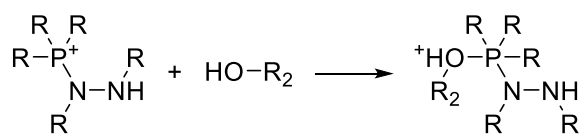

4. Bond cleavage and proton exchange

   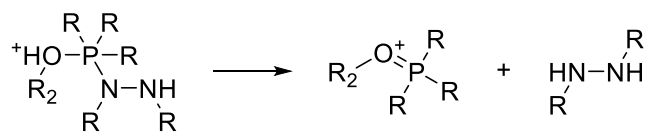

5. SN2 type reaction

   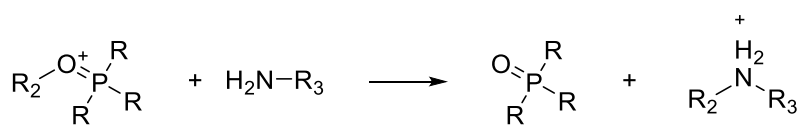

6. Deprotonation of protonated amine

   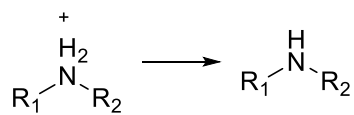

## II. React with SOCl2 or POCl3

Required agent: 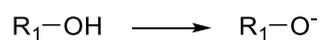

Elementary reaction

1. Deprotonation of alcohol

    R₁−OH ⟶ R₁−O⁻

2. Addition of alkoxide

    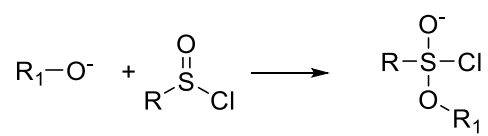

3. Chloride leaves

    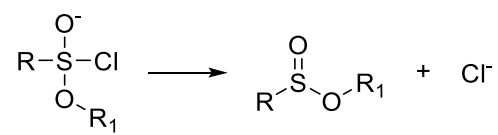

4. Nucleophilic substitution

    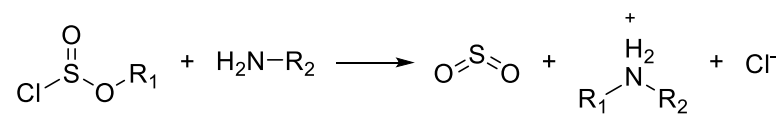

5. Deprotonation of protonated amine

    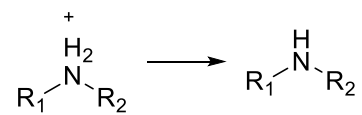

# Reaction class name: Friedel-Crafts acylation

## I. Mitsunobu style reaction

Required agent: AlCl₃

Elementary reaction

1. Lewis acid activation

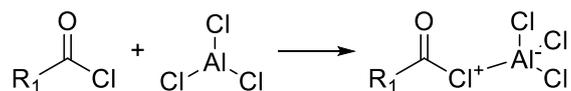

2. AlCl4 leaves

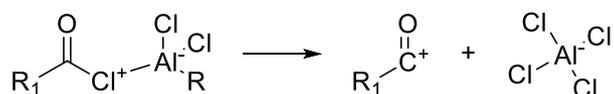

3. Addition to aromatic ring

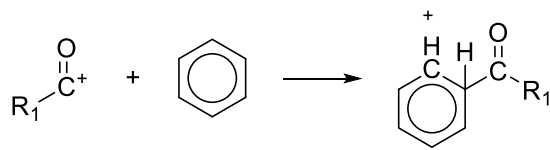

4. Proton exchange

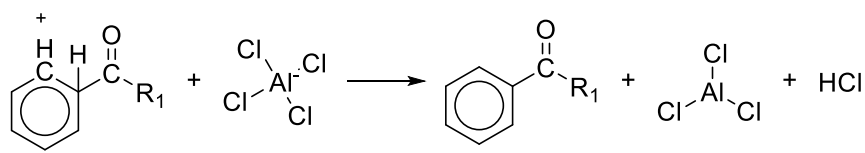

# Reaction class name: Bromo Sonogashira coupling / Iodo Sonogashira coupling / Chloro Sonogashira coupling / Triflyloxy Sonogashira coupling / Bromo Stille reaction / Chloro Stille reaction / Iodo Stille reaction

## I. Reaction with element Pd

Required agent: Pd

Elementary reaction

1. Copper activation

   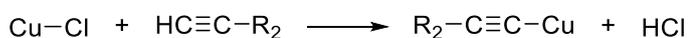

2. Oxidative addition

   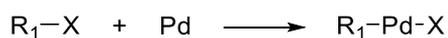

3. Transmetalation

   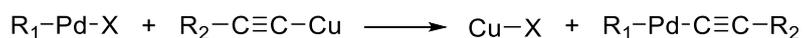

   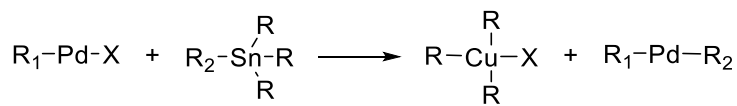

4. Reductive elimination

   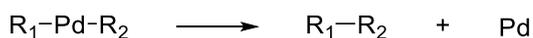

## II. Reaction with PdCl2

Required agent: PdCl$_2$ and P

Elementary reaction

1. Palladium-ligand formation

   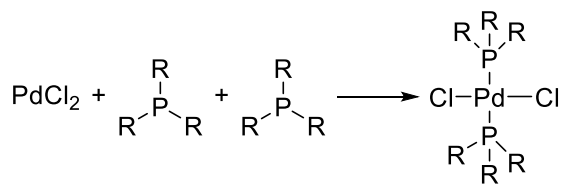

2. Palladium reduction

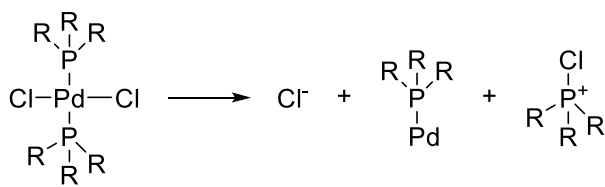

3. Copper activation

   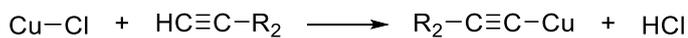

4. Oxidative addition

   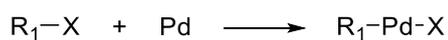

5. Transmetalation

   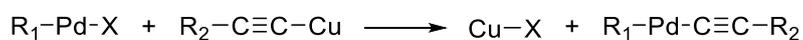

   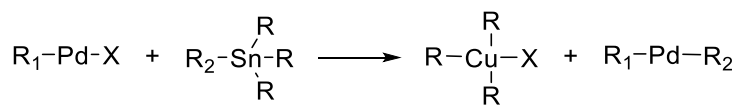

6. Reductive elimination

   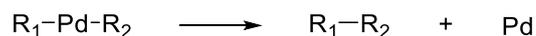

## III. Reaction with Pd coordinated with 3 or 4 ligands

Required agent: 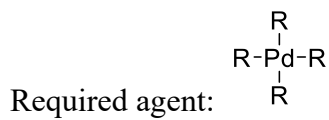

Elementary reaction

1. Copper activation

   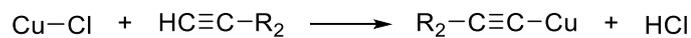

2. Oxidative addition

   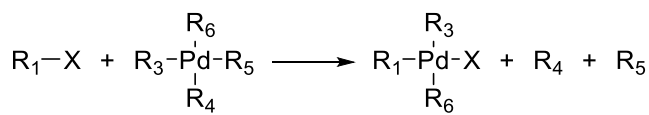

3. Transmetalation

   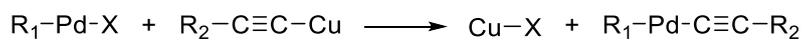

   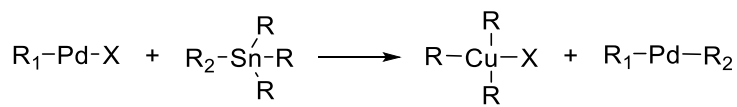

4. Reductive elimination

$$R_1-Pd-R_2 \longrightarrow R_1-R_2 + Pd$$

## IV. Reaction with Pd coordinated with 2 ligands

Required agent: $R_3-Pd-R_2$

Elementary reaction

1. Copper activation

$$Cu-Cl + HC\equiv C-R_2 \longrightarrow R_2-C\equiv C-Cu + HCl$$

2. Oxidative addition

$$R_1-X + R_3-Pd-R_4 \longrightarrow R_1-\underset{R_4}{\overset{R_3}{Pd}}-X$$

3. Transmetalation

$$R_1-Pd-X + R_2-C\equiv C-Cu \longrightarrow Cu-X + R_1-Pd-C\equiv C-R_2$$

$$R_1-Pd-X + R_2-\underset{R}{\overset{R}{Sn}}-R \longrightarrow R-\underset{R}{\overset{R}{Cu}}-X + R_1-Pd-R_2$$

4. Reductive elimination

$$R_1-Pd-R_2 \longrightarrow R_1-R_2 + Pd$$

# Reaction class name: Knoevenagel condensation / Claisen-Schmidt condensation

## I. Reaction

Required agent: None

Elementary reaction

1. Deprotonation of alpha position carbon

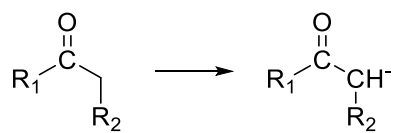

2. Addition of enolate

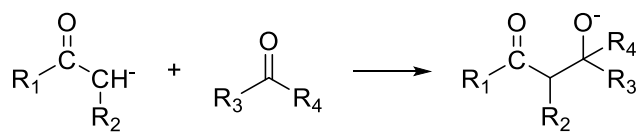

3. Protonation of alkoxide

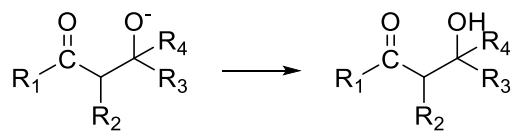

4. Condensation

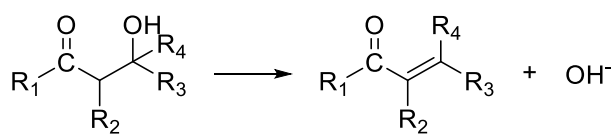

# Reaction class name: Wohl-Ziegler bromination

**I. Reaction with radical**

Required agent: 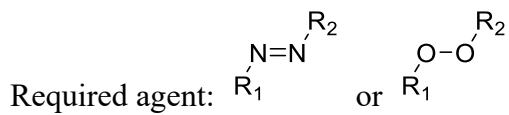

Elementary reaction

1. Radical formation

   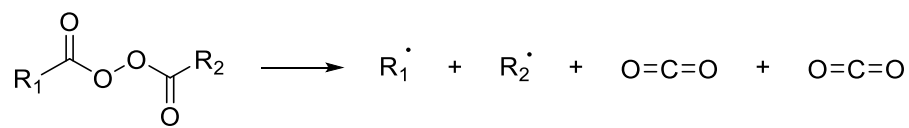

2. Reaction with NBS

   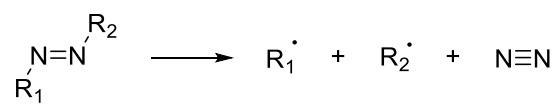

3. Propagation

   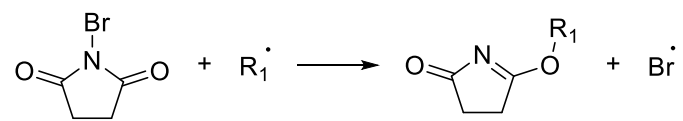

   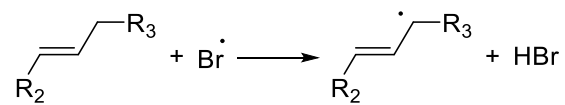

4. Termination

   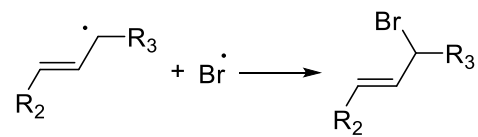

**II. Reaction with NBS**

Required agent: None

Elementary reaction

1. Radical formation

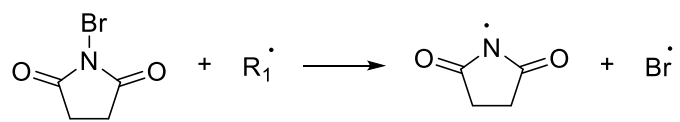

2. Propagation

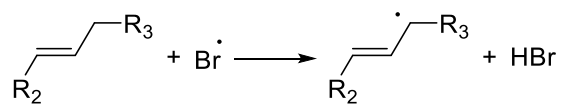

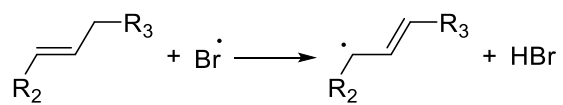

3. Termination

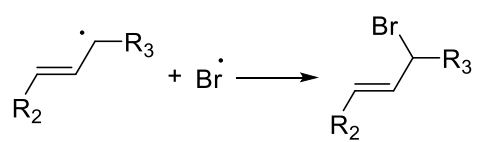

**Reaction class name: Bromo Grignard reaction / Bromo Grignard + ester reaction / Chloro Grignard reaction / Chloro Grignard + ester reaction / Grignard ester substitution / Iodo Grignard reaction / Grignard + acid chloride ketone synthesis**

**I. Reaction**

Required agent: None

Elementary reaction

1. Addition

    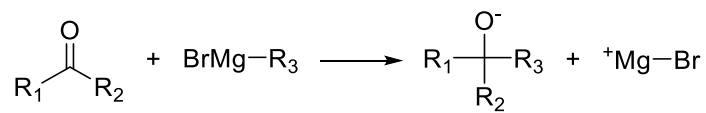

2. Leaving or workup

    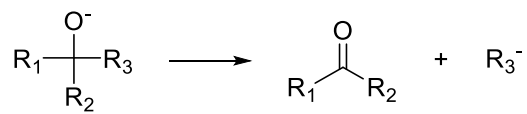

    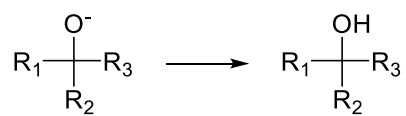

# Reaction class name: Bromo Grignard preparation / Iodo Grignard preparation / Chloro Grignard preparation

**I. Reaction**

Required agent: None

Elementary reaction

1. Radical formation

    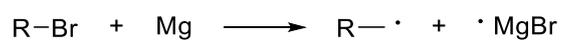
    R−Br + Mg ⟶ R—· + ·MgBr

2. Termination

    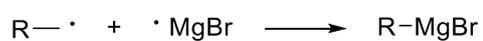
    R—· + ·MgBr ⟶ R−MgBr

# Reaction class name: Wittig olefination

## I. Reaction

Required agent: None

Elementary reaction

1. Cycloaddition

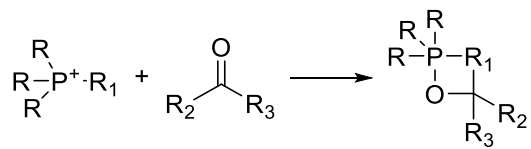

2. Cyclo-reversion

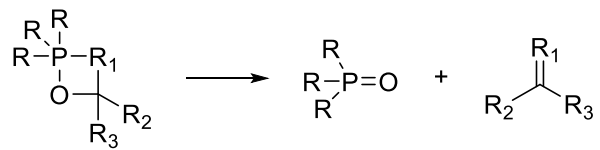

# Reaction class name: Darzens chlorination / Darzens bromination

**I. Reaction**

Required agent: None

Elementary reaction

1. Deprotonation of alcohol

   R₁−OH ⟶ R₁−O⁻

2. Addition

   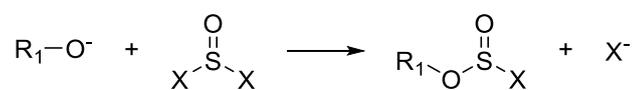

3. Nucleophilic substitution

   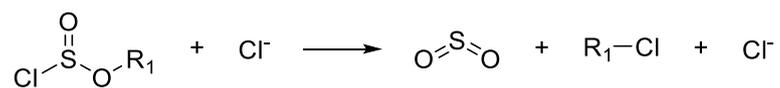

# Reaction class name: Weinreb amide synthesis

## I. Reaction

Required agent: None

Elementary reaction

1. Amine addition

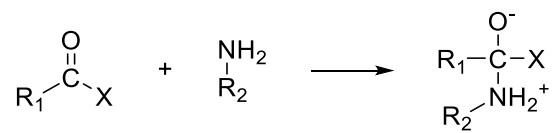

2. Leaving group leaves

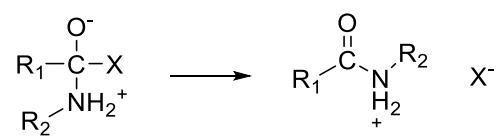

3. Deprotonation of amide

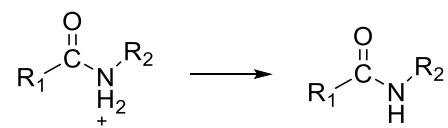

# Reaction class name: Weinreb ketone synthesis

## I. Reaction

Required agent: None

Elementary reaction

1. Nucleophilic addition

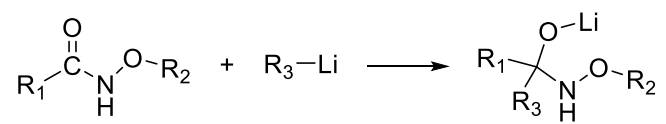

2. Ketone formation

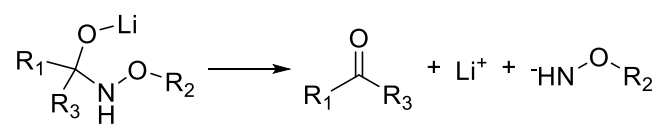

3. Protonation of amide anion

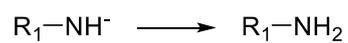

# Reaction class name: Azide-alkyne Huisgen cycloaddition

**I. Reaction**

Required agent: None

Elementary reaction

1. Cycloaddition

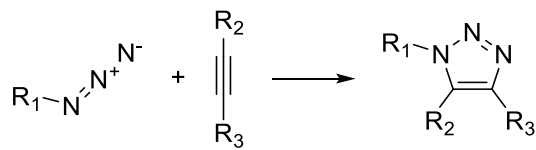

# Reaction class name: Aldehyde Dess-Martin oxidation / Ketone Dess-Martin oxidation

**I. Reaction**

Required agent: None

Elementary reaction

1. Substitution

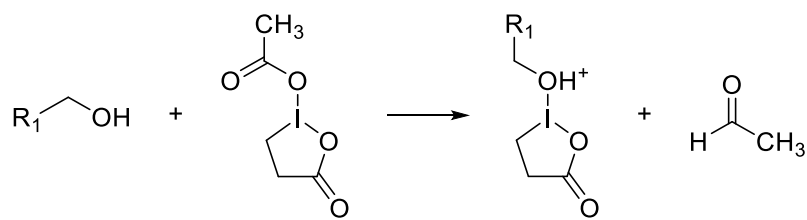

2. Deprotonation

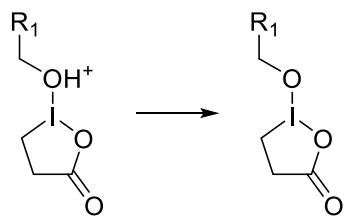

3. Oxidation

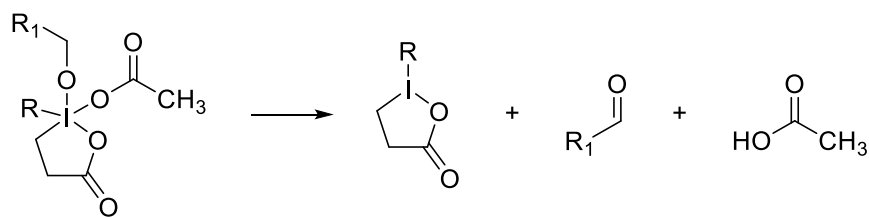

# Reaction class name: Pinner reaction

## I. Reaction

Required agent: None

Elementary reaction

1. Protonation of nitrile

    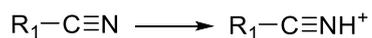

    R₁—C≡N ⟶ R₁—C≡NH⁺

2. Addition of alcohol or amine

    Acidic condition p$K_a$ < -10

    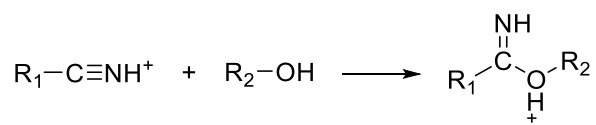

    Relatively milder condition

    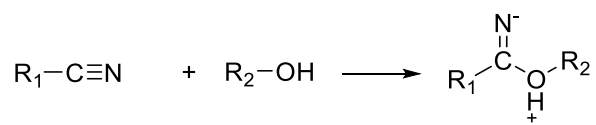

3. Deprotonation or proton exchange

    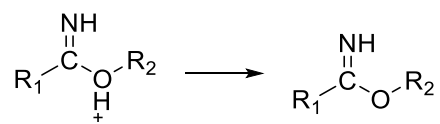

    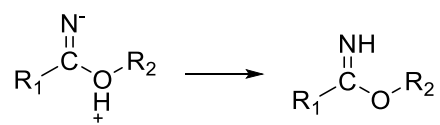

# Reaction class name: Amide Schotten-Baumann / Sulfonamide Schotten-Baumann / Sulfonic ester Schotten-Baumann / Carbamate Schotten-Baumann / Ester Schotten-Baumann / N-Cbz protection

## I. Reaction

Required agent: None

Elementary reaction

1. Amine addition

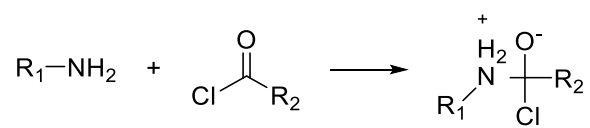

2. Proton exchange

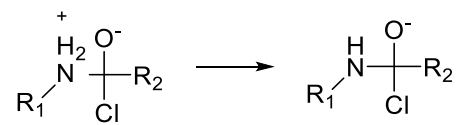

3. Halide leaves

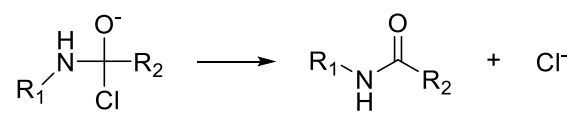

# Reaction class name: Mesyloxy N-alkylation

## I. Reaction

Required agent: None

Elementary reaction

1. Substitution

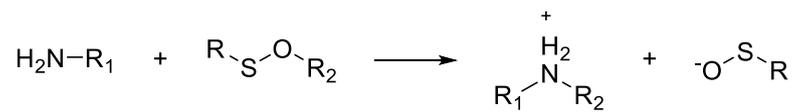

2. Deprotonation

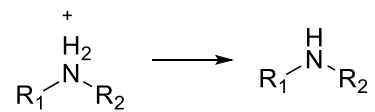

# Reaction class name: Keto alpha-alkylation

## I. Reaction

Required agent: None

Elementary reaction

1. Deprotonation of alpha position carbon

    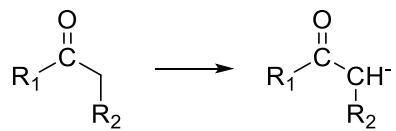

2. Substitution

    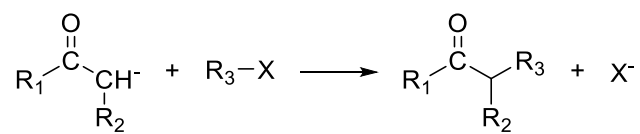

# Reaction class name: CO2H-Me deprotection / CO2H-Et deprotection / CO2H-tBu deprotection / Ester hydrolysis

## I. Deprotection with OH-

Required agent: OH⁻

Elementary reaction

1. Hydroxide addition

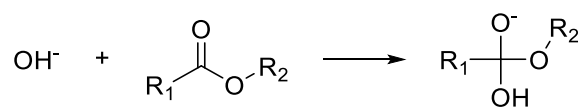

2. Alkoxide leaves

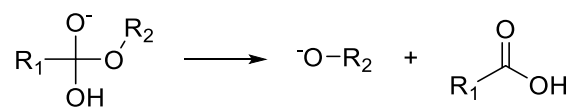

## II. Deprotection with Alkali hydroxide

Required agent: NaOH

Elementary reaction

1. Hydroxide addition

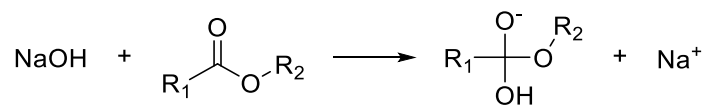

2. Alkoxide leaves

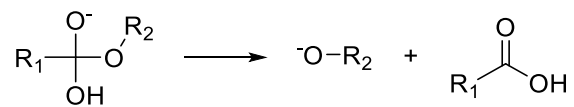

## II. Deprotection with water

Required agent: H₂O

Elementary reaction

1. Water addition

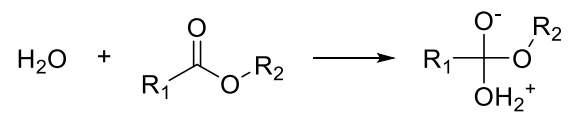

2. Deprotonation

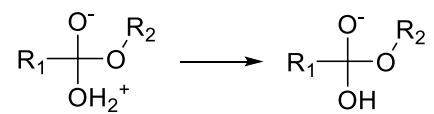

3. Alkoxide leaves

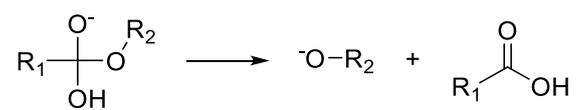